\documentclass{article}

\usepackage{microtype}
\usepackage{graphicx}
\usepackage{subfig}
\usepackage{booktabs} 

\usepackage{multirow}
\usepackage{float}
\usepackage{bm}

\usepackage{amsmath}

\allowdisplaybreaks[4]
\usepackage{amssymb}

\usepackage{hyperref}

\usepackage[accepted]{icml2021}

\icmltitlerunning{Controllable Pareto Multi-Task Learning}

\DeclareMathOperator*{\argmin}{arg\,min}

\def\ddefloop#1{\ifx\ddefloop#1\else\ddef{#1}\expandafter\ddefloop\fi}

\newcommand\norm[1]{||#1||}

\def\ddef#1{\expandafter\def\csname v#1\endcsname{\ensuremath{\boldsymbol{#1}}}}
\ddefloop ABCDEFGHIJKLMNOPQRSTUVWXYZabcdefghijklmnopqrstuvwxyz\ddefloop

\def\ddef#1{\expandafter\def\csname v#1\endcsname{\ensuremath{\boldsymbol{\csname #1\endcsname}}}}
\ddefloop {alpha}{beta}{gamma}{delta}{epsilon}{varepsilon}{zeta}{eta}{theta}{vartheta}{iota}{kappa}{lambda}{mu}{nu}{xi}{pi}{varpi}{rho}{varrho}{sigma}{varsigma}{tau}{upsilon}{phi}{varphi}{chi}{psi}{omega}{Gamma}{Delta}{Theta}{Lambda}{Xi}{Pi}{Sigma}{varSigma}{Upsilon}{Phi}{Psi}{Omega}{ell}\ddefloop

\def\ddef#1{\expandafter\def\csname bb#1\endcsname{\ensuremath{\mathbb{#1}}}}
\ddefloop ABCDEFGHIJKLMNOPQRSTUVWXYZ\ddefloop

\begin{document}

\twocolumn[
\icmltitle{Controllable Pareto Multi-Task Learning}



\icmlsetsymbol{equal}{*}

\begin{icmlauthorlist}
\icmlauthor{Xi Lin}{to}
\icmlauthor{Zhiyuan Yang}{to}
\icmlauthor{Qingfu Zhang}{to}
\icmlauthor{Sam Kwong}{to}
\end{icmlauthorlist}

\icmlaffiliation{to}{Department of Computer Science, City University of Hong Kong, Hong Kong}

\icmlcorrespondingauthor{Xi Lin}{xi.lin@my.cityu.edu.hk}

\icmlkeywords{Multi-Task Learning, Multi-Objective Optimization}

\vskip 0.3in
]



\printAffiliationsAndNotice{}  

\begin{abstract}
A multi-task learning (MTL) system aims at solving multiple related tasks at the same time.  With a fixed model capacity, the tasks would be conflicted with each other, and the system usually has to make a trade-off among learning all of them together. For many real-world applications where the trade-off has to be made online, multiple models with different preferences over tasks have to be trained and stored. This work proposes a novel controllable Pareto multi-task learning framework, to enable the system to make real-time trade-off control among different tasks with a single model. To be specific, we formulate the MTL as a preference-conditioned multiobjective optimization problem, with a parametric mapping from preferences to the corresponding trade-off solutions. A single hypernetwork-based multi-task neural network is built to learn all tasks with different trade-off preferences among them, where the hypernetwork generates the model parameters conditioned on the preference. For inference, MTL practitioners can easily control the model performance based on different trade-off preferences in real-time. Experiments on different applications demonstrate that the proposed model is efficient for solving various MTL problems.
\end{abstract}

\section{Introduction}
\label{sec_intro}

\begin{figure}[t]
    \centering
    \includegraphics[width=0.8 \linewidth]{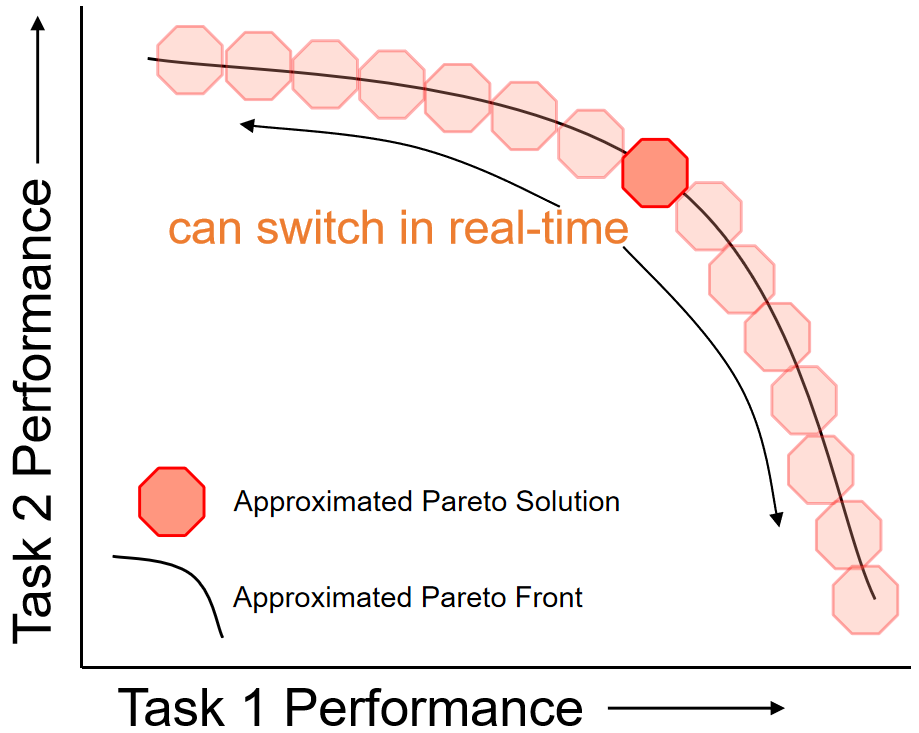}
    \caption{\textbf{Controllable Pareto MTL} allows practitioners to control the trade-offs among different tasks in real time with a single model, which could be desirable for many MTL applications.}
    \label{Controllable_MOPMTL}
\end{figure}

Multi-task learning (MTL) is important for many real-world applications, such as in computer vision~\citep{kokkinos2017ubernet}, natural language processing~\citep{subramanian2018learning}, and reinforcement learning~\citep{van2014multi}.  In these problems, multiple tasks are needed to be learned at the same time. An MTL system usually builds a single model to learn several related tasks together, in which the positive knowledge transfer could improve the performance for each task. In addition, using one model to conduct multiple tasks is also good for saving storage costs and reducing the inference time, which could be crucial for many applications~\citep{standley2020which}.

However, with a fixed learning capacity, different tasks could be conflicted with each other, and can not be optimized simultaneously~\citep{zamir2018taskonomy}. The practitioners might need to carefully assign the tasks into different groups to achieve the best performance~\citep{standley2020which}.  A considerable effort is also needed to find a suitable way to balance the performance of each task~\citep{kendall2017multi,chen2018grad,sener2018multi}. Recently, a few methods have been proposed to train and store multiple models with different trade-offs among the tasks~\citep{lin2019pareto,mahapatramulti2020multi,ma2020efficient}.

In many real-world MTL applications, the system needs to make a trade-off among different tasks in real time, and it is desirable to have the whole set of optimal trade-off solutions. For example, in a self-driving system, multiple tasks must be conducted simultaneously but also compete for a fixed resource (e.g., fixed total inference time threshold), and their preferences could change in real time for different scenarios~\citep{karpathy2019multi}. A recommendation system needs to balance multiple criteria among different stakeholders simultaneously, and making trade-off adjustment would be a crucial component~\citep{milojkovic2019multi}. Consider the huge storage cost, it is far from ideal to train and store multiple models to cover different trade-off preferences, which is also not good for real-time adjustment.  

In this paper, we propose a novel controllable Pareto multi-task learning framework, to learn the whole trade-off curve for all tasks with a single model. As shown in Fig.\ref{Controllable_MOPMTL}, at the inference time, MTL practitioners can easily control the trade-off among tasks based on their preferences. The main contributions of this work are:

\begin{itemize}
    \item We formulate solving an MTL problem as a preference-conditioned multiobjective optimization problem, and propose a novel solution generator to learn the whole trade-off curve for the given problem. 
    \item We propose a general hypernetwork-based multi-task neural network framework, and develop an efficient end-to-end algorithm to simultaneously optimize different trade-off preferences via a single model.
    \item Experiments on different MTL problems validate that the proposed method can successfully learn the trade-off curves and support real-time trade-off control. 
\end{itemize}

\section{Related Work}
\label{sec_related_work}

\textbf{Multi-Task Learning.} The current works on deep multi-task learning mainly focus on designing novel network architecture and constructing efficient shared representation among tasks~\citep{zhang2017survey,ruder2017overview}. Different deep MTL networks, with hard or soft parameters sharing structures, haven been proposed in the past few years~\citep{misra2016cross,long2017learning,yang2017deep}. However, how to properly combine and learn different tasks together remains a basic but challenging problem for MTL applications. Although it has been proposed for more than two decades, the simple linear tasks scalarization approach is still the current default practice to combine and train different tasks in MTL problems~\citep{caruana1997multitask}. 

Some adaptive weight methods have been proposed to better combine all tasks in MTL problems with a single model~\citep{kendall2017multi,chen2018grad,liu2019end,yu2020gradient}. However, analysis on the relations among tasks in transfer learning~\citep{zamir2018taskonomy} and multi-task learning~\citep{standley2020which} show that some tasks might conflict with each other and can not be optimized at the same time. Sener and Koltun~\citep{sener2018multi} propose to treat MTL as a multiobjective optimization problem, and find a single Pareto stationary solution among different tasks. Pareto MTL~\citep{lin2019pareto} generalizes this idea, and proposes to find a set of solutions with different trade-off preferences. Recent works focuses on generating diverse and dense Pareto stationary solutions~\citep{mahapatramulti2020multi,ma2020efficient}. These methods need to train and store multiple models to cover the trade-off curve for a given problem, which is undesirable for many real-world applications.

Very recently, there is a concurrent work~\citep{navon2021learning} that also independently proposes to learn the entire trade-off curve for MTL problems by hypernetwork. Their work emphasizes the runtime efficiency on training for multiple preferences, and validate their models on small scale problems. We highlight our method's advantage on supporting real-time preference control for inference, and show it can scale well for large scale MTL model. 

\textbf{Multi-Objective Optimization.} Multi-Objective optimization itself is a popular research topic in the optimization community. Many gradient-based and gradient-free algorithms have been proposed in the past decades~\citep{fliege2000steepest,desideri2012mutiple, miettinen2012nonlinear}. The closest method to our approach is the decomposition-based multi-objective evolutionary algorithm (MOEA/D~\citep{zhang2007moea}), which decomposes a multi-objective optimization problem into finite preference-based subproblems and solves them at the same time. We propose to learn the whole trade-off curve for a given problem with a single model, which might contain infinite trade-off solutions.

In addition to MTL, multi-objective optimization algorithms also can be used in reinforcement learning~\citep{van2014multi} and neural architecture search (NAS)~\citep{elsken2018efficient,lu2020neural}. However, most methods directly use or modify well-studied multi-objective algorithms to find a single solution or a finite number of Pareto solutions, and do not support real-time trade-off control. \citet{parisi2016multi} proposed to learn the Pareto manifold for a multi-objective reinforcement learning problem. However, since this method does not consider the preference for generation, it does not support real-time preference adjustment. Recently, some methods have been proposed to learn preference-based solution adjustment for multi-objective reinforcement learning~\citep{yang2019generalized} and image generation~\citep{Dosovitskiy2020You} with simple linear combinations and model adaptions. This paper uses a hypernetwork to generate all parameters for the main multi-task neural network conditioned on different preferences.

\textbf{HyperNetworks.} The hypernetwork is initially proposed for dynamic modeling and model compression~\citep{schmidhuber1992learning, ha2017hypernetworks}. It also leads to various novel applications such as hyperparameter optimization~\citep{brock2018smash,mackay2019self}, Bayesian inference~\citep{krueger2018bayesian,dwaracherla2020hypermodels}, and transfer learning~\citep{oswald2020continual, meyerson2019modular}. Recently, some discussions have been made on its initialization method~\citep{chang2020principled}, the optimization dynamic~\citep{littwin2020on}, and its relation to other multiplicative interaction methods~\citep{jayakumar2020multiplicative}.

\begin{figure*}[t]
    \centering
    \includegraphics[width= 0.85 \linewidth]{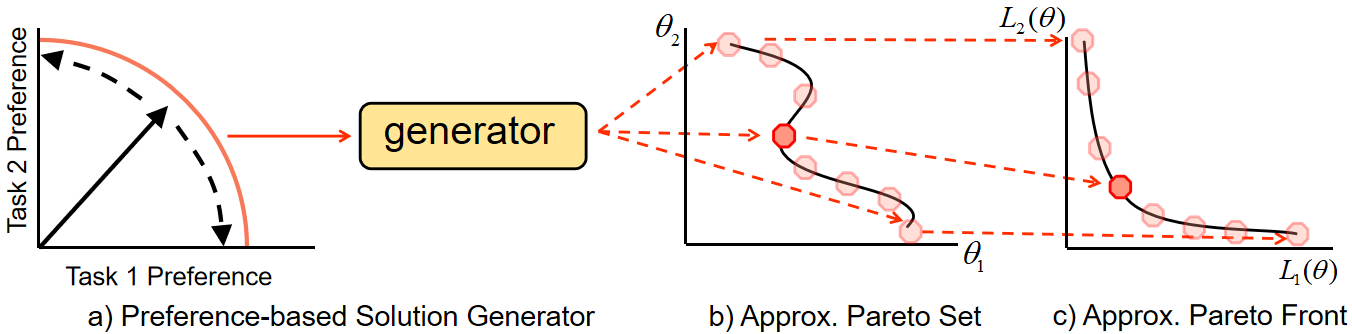}
    \caption{\textbf{Controllable Pareto MTL} can directly generate a corresponding solution based on a given trade-off preference among tasks. (a) A practitioner controls the $m$-dimensional preference vector $\vp$ to specify a trade-off preference among $m$ tasks. (b) A trained preference-based generator directly generates a solution $\vtheta_{\vp}$ conditioned on $\vp$. (c) The generated solution $\vtheta_{\vp}$ should have objective values $L(\vtheta_{\vp})$ that satisfies the trade-off preference among $m$ tasks. The number of possible preference vectors and the corresponding solutions can be infinite for an MTL problem. We develop the idea in Section \ref{pareto_generator} and propose the MTL model in Section \ref{sec_framework}.}
    \label{controllable_pareto_mtl}
\end{figure*}

\section{MTL as Multi-Objective Optimization}
\label{sec_pareto_mtl}

An MTL problem involves learning multiple related tasks at the same time. For training a deep multi-task neural network, it is to minimize the losses for multiple tasks:
\begin{align}\label{eq:mop}
\mathrm{min}_{\theta} \ \mathcal{L}(\theta) = (\mathcal{L}_1(\theta),\mathcal{L}_2(\theta),\cdots,
\mathcal{L}_m(\theta)),
\end{align}
where $\theta$ is the neural network parameters and $\mathcal{L}_i(\theta)$ is the empirical loss of the $i$-th task. For learning all $m$ tasks together, an MTL system usually aims to minimize all losses at the same time. However, in many problems, it is impossible to find a single best solution to optimize all the losses simultaneously. With different trade-offs among the tasks, the problem~(\ref{eq:mop}) could have a set of Pareto solutions which satisfy the following definitions~\citep{zitzler1999multiobjective}:

\textbf{Pareto dominance.} Let $\theta_a,\theta_b$ be two solutions for problem~(\ref{eq:mop}), $\theta_a$ is said to dominate $\theta_b$ ($\theta_a \prec \theta_b$) if and only if $\mathcal{L}_i(\theta_a) \leq \mathcal{L}_i(\theta_b), \forall i \in \{1,...,m\}$ and $\mathcal{L}_j(\theta_a) < \mathcal{L}_j(\theta_b), \exists j \in \{1,...,m\}$.

\textbf{Pareto optimality.} $\theta^{\ast}$ is a Pareto optimal solution if there does not exist $\hat \theta$ such that $\hat \theta \prec \theta^{\ast}$.
The set of all Pareto optimal solutions is called the Pareto set. The image of the Pareto set in the objective space is called the Pareto front.

\textbf{Finite Solutions Approximation.} The number of Pareto solutions could be infinite, and their objective values are on the boundary of the valid value region~\citep{boyd2004convex}. Under mild conditions, the whole Pareto set and Pareto front would be $(m - 1)$-dimensional manifolds in the solution space and objective space, respectively~\citep{miettinen2012nonlinear}. For a general multiobjective optimization problem, no method can guarantee to find the Pareto front. Traditional multiobjective optimization algorithms aim at finding a set of finite solutions to approximate the whole Pareto set and Pareto front:
\begin{eqnarray}
	\hat S = \{ \hat \vtheta_1, \cdots, \hat \vtheta_K \}, \hat F = \{\mathcal{L}(\hat \vtheta_1),\cdots,\mathcal{L}(\hat \vtheta_K)\},
\end{eqnarray}
where $\hat S$ is the approximated Pareto set of $K$ estimated solutions, and $\hat F$ is the set of corresponding objective vectors in the objective space. The current multiobjective optimization based MTL methods aim to find a single ($K = 1$) \citep{sener2018multi} or a set of multiple ($K > 1$) Pareto stationary solutions~\citep{lin2019pareto, mahapatramulti2020multi,ma2020efficient} for a given problem. For an MTL problem, each solution is a large deep neural network. To well cover a trade-off curve, the number of required solutions might grow exponentially with the number of tasks~\cite{lin2019pareto,ma2020efficient}. The huge training and storage cost make these methods less practical for real-world applications. 

\section{Preference-Based Solution Generator}
\label{pareto_generator}

Instead of training multiple models, we propose to directly learn the whole trade-off curve for a MTL problem with a single model. Similar to other multi-objective optimization algorithms, our method can not guarantee to find the ground truth Pareto front, but we show that it can find a good approximated trade-off curve for various problems.

As shown in Fig.~\ref{controllable_pareto_mtl}, we want to build a solution generator to map a preference vector $\vp$ to its corresponding solution $\vtheta_{\vp}$. If an optimal generator $\vtheta_{\vp} = g(\vp|\vphi^*)$ is obtained, MTL practitioners can assign their preference via the preference vector $\vp$, and directly obtain the corresponding solution $\vtheta_{\vp}$ with the specific trade-off among tasks. With the solution generator, we can obtain the approximated Pareto set/front:
\begin{eqnarray}
	\hat S = \{\vtheta_{\vp} = g(\vp|\vphi^*) | \vp \in \vP \}, \hat F = \{\mathcal{L}(\vtheta_{\vp})| \vtheta_{\vp} \in \hat S \},
\end{eqnarray}
where $\vP$ is the set of all valid preference vectors, $g(\vp|\vphi^*)$ is the solution generator with optimal parameters $\phi*$. Once we have a proper generator $g(\vp|\vphi^*)$, we can reconstruct the whole approximated Pareto set $\hat S$ and the approximated Pareto front $\hat F$ by going through all possible preference vector $\vp$. In the rest of this section, we discuss two approaches to define the form of preference vector $\vp \in \vP$ and its connection to the corresponding solution $\vtheta_p = g(\vp|\vphi^*)$. 

\textbf{Preference-Based Linear Scalarization:} A simple and straightforward approach is to define the preference vector $\vp$ and the corresponding solution $\vtheta_{\vp}$ via the weighted linear scalarization: 
\begin{eqnarray}\label{eq_linear}
\vtheta_{\vp} = g(\vp|\vphi^*) = \argmin_{\vtheta} \sum_{i=1}^{m}\vp_i \mathcal{L}_i(\vtheta),
\end{eqnarray}
where the preference vector $\vp = (\vp_1, \vp_2, \cdots, \vp_m)$ is the weight for each task, and $\vtheta_{\vp}$ is the optimal solution for the weighted linear scalarization. If we further require $\sum \vp_i = 1$, the set of all valid preference vector $\vP$ is an $(m-1)$-dimensional manifold in $\bbR^m$, while the Pareto set and Pareto front are both $(m-1)$-dimensional manifolds. It should be noticed that the solution for the right-hand side of equation~(\ref{eq_linear}) might be not unique. For simplicity, we assume there is always a one-to-one mapping in this paper.  

Although this approach is straightforward, it is not optimal for multiobjective optimization. Linear scalarization can not find any Pareto solution on the non-convex part of the Pareto front~\citep{das1997a,boyd2004convex}. In other words, unless the problem has a convex Pareto front, the generator defined by linear scalarization cannot cover the whole Pareto set manifold. 

\begin{figure}[t]
\centering
\includegraphics[width= 0.70 \linewidth]{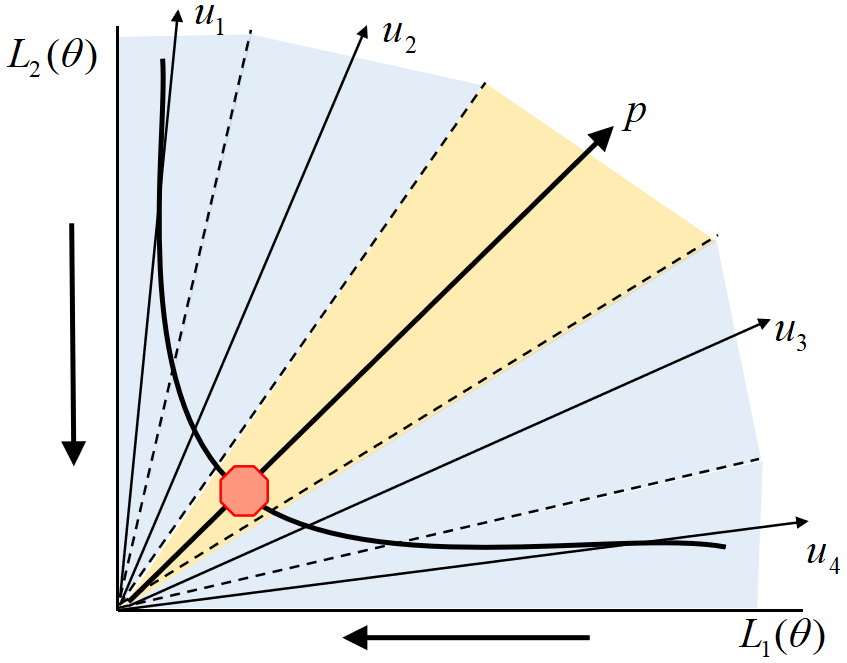}
    	\caption{An MTL problem is decomposed by a set of unit vectors. }
        \label{MOPM2M}
\end{figure}

\textbf{Preference-Based Multiobjective Optimization:} To better approximate the Pareto set, we generalize the idea of decomposition-based multiobjective optimization~\citep{zhang2007moea,liu2014decomposition} and Pareto MTL~\citep{lin2019pareto} to connect the preference vector and the corresponding Pareto solution. To be specific, we define the preference vector $\vp$ as an $m$-dimensional unit vector in the loss space, and the corresponding solution $\vtheta_{\vp}$ is the one on the Pareto front which has the smallest angle with $\vp$. 

The idea is illustrated in in Fig.~\ref{MOPM2M}. With a set of randomly generated unit reference vectors $\vU = \{\vu^{(1)}, \cdots, \vu^{(K)}\}$ and the preference vector $\vp$, an MTL problem is decomposed into different regions in the loss space. We call the region closest to $\vp$ as its preferred region. The corresponding Pareto solution $\vtheta_{\vp}$ is the solution that belongs to the preferred region and on the Pareto front. Formally, we can define the corresponding Pareto solution as:
\begin{eqnarray}\label{eq_pareto_mtl}
\begin{aligned}
\vtheta_{\vp} = g(\vp|\vphi^*) = \argmin_{\vtheta} \mathcal{L}(\vtheta), \text{s.t.}\mathcal{L}(\vtheta) \in \Omega(\vp,\vU), 
\end{aligned}
\end{eqnarray}
where $\mathcal{L}(\vtheta)$ is the loss vector and $\Omega(\vp,\vU)$ is the constrained preference region conditioned on the preference and reference vectors $\Omega(\vp, \vU) = \{\vv \in R^m_{+}| \angle(\vv,\vp) \leq \angle(\vv,\vu^{(j)}), \forall j = 1,...,K\}$.

The constraint is satisfied if the loss vector $\mathcal{L}(\vtheta)$ has the smallest angle with the preference vector $\vp$. The corresponding solution $\vtheta_{\vp}$ is restricted Pareto optimal in $\Omega(\vp,\vU)$. Since we require the preference vectors should be unit vector $||\vp||^2 = 1$ in the $m$-dimensional space, the set of all valid preference vectors $\vP$ is an $(m-1)$-dimensional manifold in $\bbR^m$. We provide more discussions in the Appendix. 

\section{Controllable Pareto Multi-Task Learning}
\label{sec_framework}

\begin{figure}[t]
\centering
\subfloat[Controllable Pareto MTL Network]{\includegraphics[width = 0.45\textwidth]{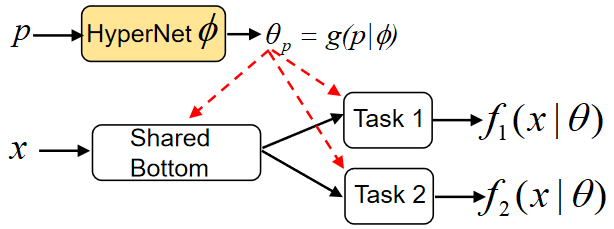}} \\
\subfloat[Chunked Hypernetwork]{\includegraphics[width = 0.40\textwidth]{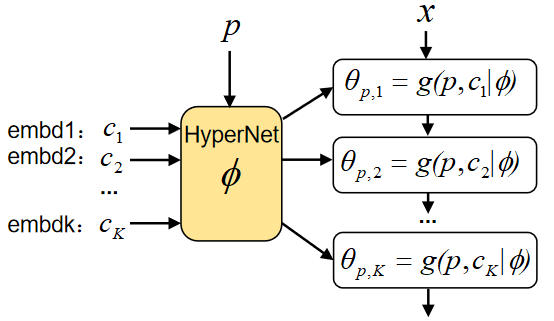}}
\caption{\textbf{The Controllable Pareto Multi-Task Network:} (a) The main MTL network has a fixed structure, and all its parameters are generated by the hypernetwork conditioned on the preference vector. (b) With the chunking method, the hypernetwork can separately generate parameters for different parts of the main network.}
\label{hypernet}
\end{figure}

In this section, we propose a hypernetwork-based multi-task neural network framework, along with an efficient end-to-end optimization procedure to solve the MTL problem.

\subsection{Hypernetwork-based Deep Multi-Task Network}

\begin{figure*}[t]
    \centering
    \includegraphics[width= 0.95 \linewidth]{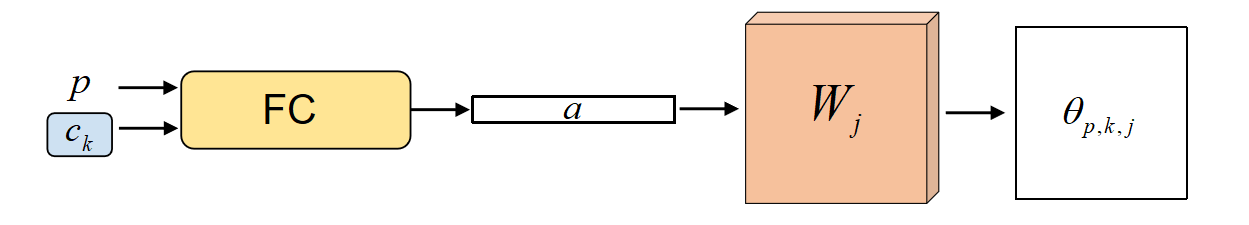}
    \caption{\textbf{Parameter Generation:} The hypernetwork takes a preference vector $\vp$ as input and generates the parameter $\theta_{p,k,j}$ for part of the main MTL network, where $\vc_k$ is the chunk embedding, $\textbf{FC}$ is the fully connected layers, $\va$ is the generated vector, and $\vW_j$ is a parameter tensor. All blocks in color contain trainable parameters for the hypernetwork. }
    \label{chunk_layer}
\end{figure*}

The proposed controllable Pareto multi-task network is shown in Fig.~\ref{hypernet}(a). As discussed in the previous section, we use a preference vector to represent a practitioner's trade-off preference among different tasks. The hypernetwork takes the preference vector  $\vp$ as its input, and generates $\vtheta_{\vp} = g(\vp|\vphi)$ as the corresponding parameters for the main MTL network. The trainable parameters to be optimized are the hypernetwork parameters $\vphi$. Practitioners can easily control the MTL network's performances on different tasks in real time, by simply adjusting the preference vector $\vp$.  

\textbf{Main MTL Model Structure:} In our proposed model, the main MTL network always has a fixed structure with a hard-shared encoder~\citep{ruder2017overview,vandenhende2020multitask} as shown in Fig.~\ref{hypernet}(a). Once the parameters $\theta_{\vp}$ are generated, an input $\vx$ to the main MTL network will first go through the shared encoder, and then all task-specific heads to obtain the outputs $f_i(\vx|\theta_p)$. Different tasks share the same encoder, and they are regularized by each other as in the traditional MTL model. Our proposed model puts the cross-tasks regularization on the hypernetwork, where it should generate a set of good encoder parameters that work well for all tasks with a given preference. 

\textbf{Hypernetwork and Scalability:} With the model compression ability powered by the hypernetwork, our proposed method can scale well to large scale models. Chunking~\citep{ha2017hypernetworks,oswald2020continual} is a commonly used method to reduce the number of parameters for the hypernetwork. As shown in Fig.~\ref{hypernet}(b), a hypernetwork can separately generate small parts of the main network $\theta_p = [\theta_{p,1},\theta_{p,2},\cdots, \theta_{p,K}]$, with a reasonable model size and multiple trainable chunk embedding $\{\vc_k\}_{k = 1}^K$, where $\theta_{p,k} = g(\vp,\vc_k|\phi)$. In this way, the hypernetwork can scale well for large MTL models. We use fully-connected networks as the hypernetwork, and the main MTL neural networks have the same structures as the models used in current MTL literature~\citep{sener2018multi, liu2019end,vandenhende2020multitask}. 

We follow the hypernetwork design proposed by \citet{ha2017hypernetworks}. The parameter generation process for a chunk of parameters is illustrated in Fig.~\ref{chunk_layer}. The hypernetwork first takes the preference vector $\vp$ and a chunk embedding $\vc_k$ as the input to a set of fully connected layers to obtain an output vector $\va_{\vp,k} \in R^d$. Then it use a linear project operation to generate the parameters $\vtheta_{\vp,k,j} = \vW_j \va$, where $\vW_j \in R^{n_{j,1} \times n_{j,2} \times d}$ contains $n_{j,1} \times n_{j,2} \times d$ trainable parameters and $\vtheta_{\vp,k,j} \in R^{n_{j,1} \times n_{j,2}}$. In the proposed hypernetwork, the fully connected layers are reusable to generate different $\va$ based on different chunk embedding. Most hypernetwork parameters are in the parameter tensors $\vW_j$.  By sharing the same $\vW$ to generate different chunks of parameters with different $\va_{\vp,k}$, the number of parameters for the hypernetwork can be significantly compressed~\citep{ha2017hypernetworks}. In this work, we keep our hypernetwork-based model to have a comparable size with the corresponding MTL model. For inference, the fully connected layers can take multiple chunk embedding to generate different $\va$ in batch, and most computation is on the linear projection. It leads to an acceptable inference latency overhead, and supports real-time control for different trade-off preferences. 

\subsection{Optimization: Learning the Generator}

Since we want to control the trade-off preference at the inference time, the proposed model should learn to perform well for all valid preference vectors rather than a single one. Suppose we have a probability distribution $P_{\vp}$ for all valid preference vector $\vp$, a general goal would be: 
\begin{eqnarray}\label{eq:sum}
 \mathrm{min}_{\vphi} \bbE_{\vp \sim P_{\vp}} \mathcal{L}(g(\vp|\vphi)).
\end{eqnarray}
However, it is hard to optimize the trainable parameters $\vphi$ within the expectation directly. We use Monte Carlo method to sample the preference vectors, and use the stochastic gradient descent algorithm to train the hypernetwork-based MTL model. We can sample one preference vector $\vp$ at a time and optimize the loss function:
\begin{eqnarray}\label{eq_sample_loss}
 \mathrm{min}_{\vphi} \mathcal{L}(g(\vp|\vphi)), \quad \vp \sim P_{\vp}.
\end{eqnarray}
At each iteration $t$, if we have a valid gradient direction $\vd_t$ for the multi-objective loss $\mathcal{L}(g(\vp|\vphi_t))$, we can simply update the parameters with gradient descent $\vphi_{t + 1} = \vphi_{t} - \eta \vd_t$. For the preference-conditioned linear scalarization case as in problem~(\ref{eq_linear}), the calculation of $\vd_t$ is straightforward:
\begin{eqnarray}\label{eq_linear_direction}
 \vd_t =  \sum_{i = 1}^{m} \vp_i \nabla_{\vphi_t} \mathcal{L}_{i}(g(\vp|\vphi_t)), \quad \vp \sim P_{\vp}.
\end{eqnarray}
For the preference-conditioned multiobjective optimization problem~(\ref{eq_pareto_mtl}), a valid descent direction should simultaneously reduce all losses and activated constraints. One valid descent direction can be written as a linear combination of all tasks with dynamic weight $\valpha_i(t)$: 
\begin{eqnarray}
    \label{update_mtl_c}
    \vd_t = \sum_{i = 1}^{m} \valpha_i(t) \nabla_{\vphi_t} \mathcal{L}_{i}(g(\vp|\vphi_t)), \vp \sim P_{\vp}, \vU \sim P_{\vU},  
\end{eqnarray}
where the coefficients $\valpha_i(t)$ is depended on both the loss functions $\mathcal{L}_{i}(g(\vp|\vphi_t))$ and the activated constraints $\mathcal{G}_j(g(\vp|\vphi_t))$. We give the detailed derivation in the Appendix due to page limit. 

\begin{figure}[t]
    \centering
    \includegraphics[width= 0.90 \linewidth]{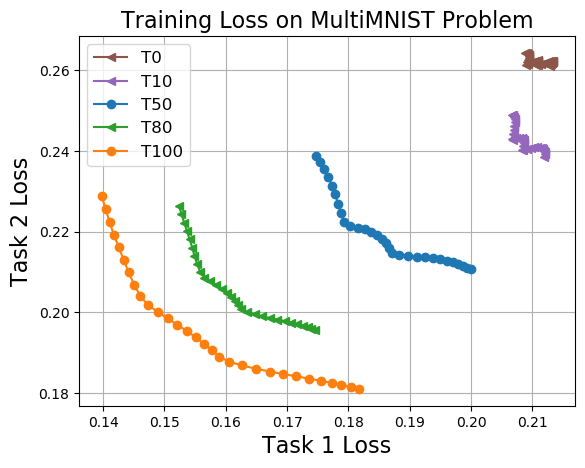}
    \caption{\textbf{The training loss trade-off curves on MultiMNIST problem at different iteration.} Our proposed algorithm continually learn the trade-off curve for all valid preferences during the optimization process. Details can be found in Section~\ref{sec_exp}.}
    \label{algorithm_and_process}
\end{figure}

\begin{algorithm}[t]
	\caption{Controllable Pareto Multi-Task Learning}
	\label{alg_framework}
 	\begin{algorithmic}[1]
    	\STATE Initialize the hypernetwork parameters $\vphi_0$
		\FOR{$t = 1$ to $T$}
		   \STATE Sample a preference vector $\vp \sim P_{\vp}$, $\vU \sim P_{\vU}$
		   \STATE Obtain a valid gradient direction $\vd_t$ from (\ref{eq_linear_direction}) or (\ref{update_mtl_c}) 
		   \STATE Update the parameters $\vphi_{t+1} = \vphi_t - \eta \vd_t$
		\ENDFOR	
		\STATE \textbf{Output:} The hypernetwork parameters $\vphi_T$
 	\end{algorithmic}
\end{algorithm}

The algorithm framework is shown in \textbf{Algorithm \ref{alg_framework}}. We use a simple uniform distribution on all valid preferences and references vector in this paper. The proposed algorithm simultaneously optimizes the hypernetwork for all valid preference vectors that represent different trade-offs among tasks. As shown in Fig.~\ref{algorithm_and_process}, it continually learns the trade-off curve during the optimization process. We obtain a preference-conditioned generator at the end, and can directly generate a solution $\vtheta_{\vp} = g(\vp|\vphi_T)$ from any preference vector $\vp$. By taking all valid preference vectors as input, we can construct the whole approximated trade-off curve. 

\section{Synthetic Multi-Objective Problem}
\label{sec_toy_example}

\begin{figure*}[t]
\centering
\subfloat[Approx. Pareto Set]{\includegraphics[width = 0.25\textwidth]{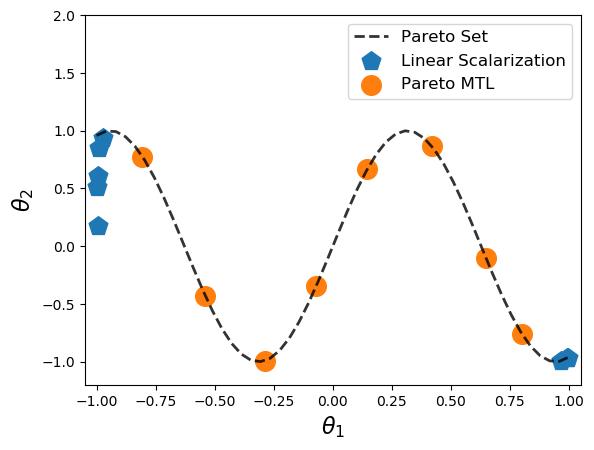}}
\subfloat[Approx. Pareto Front]{\includegraphics[width = 0.25\textwidth]{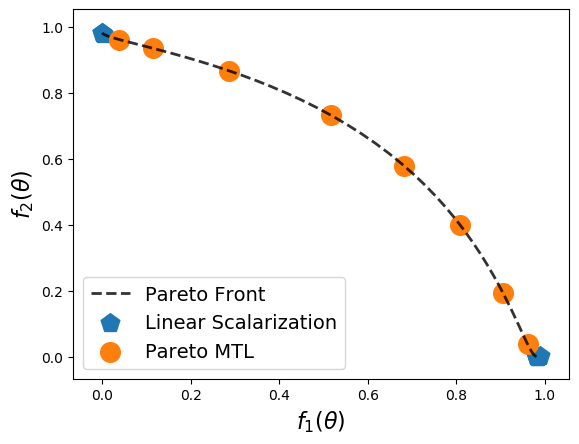}}
\subfloat[Generated Pareto Set]{\includegraphics[width = 0.25\textwidth]{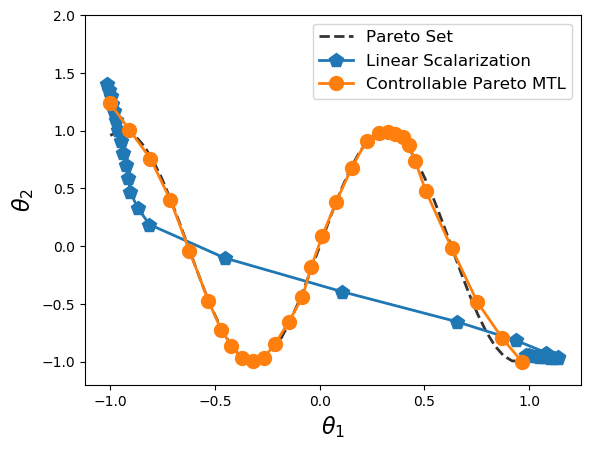}}
\subfloat[Generated Pareto Front]{\includegraphics[width = 0.25\textwidth]{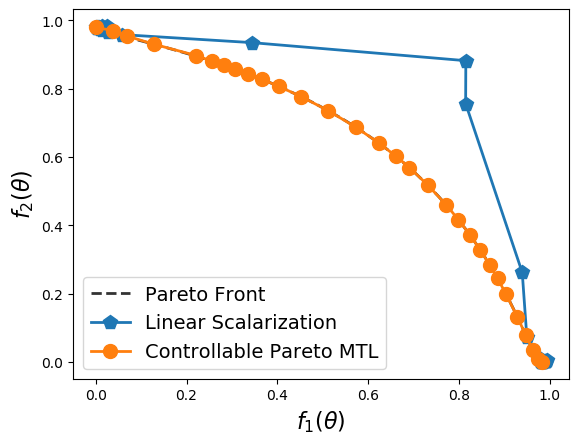}}
\caption{\textbf{Finite Solutions Approximation and Generated Pareto Front.} \textbf{(a) \& (b):} Traditional multiobjective optimization algorithms can only find a set of finite solutions to approximate the Pareto set and Pareto front. In addition, the simple linear scalarization method can not cover most part of the Pareto set/front when the ground truth Pareto front is a concave curve. \textbf{(c) \& (d):} Our proposed controllable Pareto MTL method can successfully generate the whole Pareto set and Pareto front from the preference vectors. In contrast, the linear scalarization method is only accurate for Pareto solutions near the endpoints.}
\label{synthetic_problem}
\end{figure*}

In this paper, we propose to use a hypernetwork to generate the parameters for an MTL network with different trade-off preferences. The MTL network parameters are in a high-dimensional decision space, and the optimization landscape is complicated with an unknown Pareto front. To better analyze the proposed algorithm's behavior and convergence performance, we first use it to learn the Pareto front for a low-dimensional multiobjective optimization problem.

The experimental results are shown in Fig.~\ref{synthetic_problem}. This synthetic problem has a sine-curve-like Pareto set in the solution space, and its Pareto front is a concave curve in the objective space. The detailed definition is given in the Appendix. Our proposed controllable Pareto MTL method can successfully learn and reconstruct the whole Pareto set/front from the preference vectors, while the traditional methods can only find a set of finite solutions. The simple preference-based linear scalarization method has poor performance for the problem with a totally concave Pareto front.

\section{Experiments}
\label{sec_exp}

\begin{figure*}[t]
\centering
\subfloat[MultiMNIST]{\includegraphics[width = 0.31\textwidth]{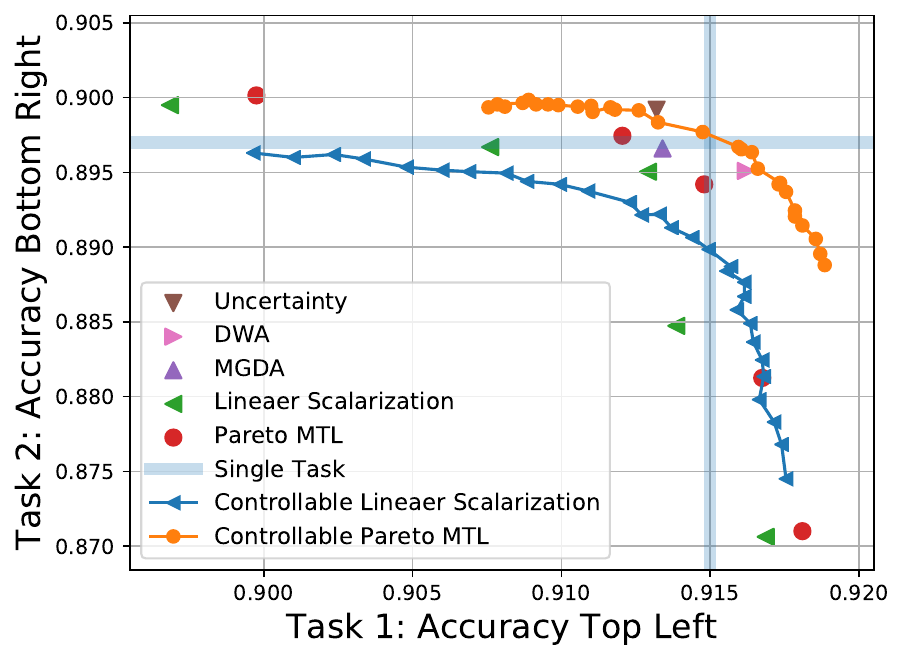}} 
\subfloat[CityScapes]{\includegraphics[width = 0.30\textwidth]{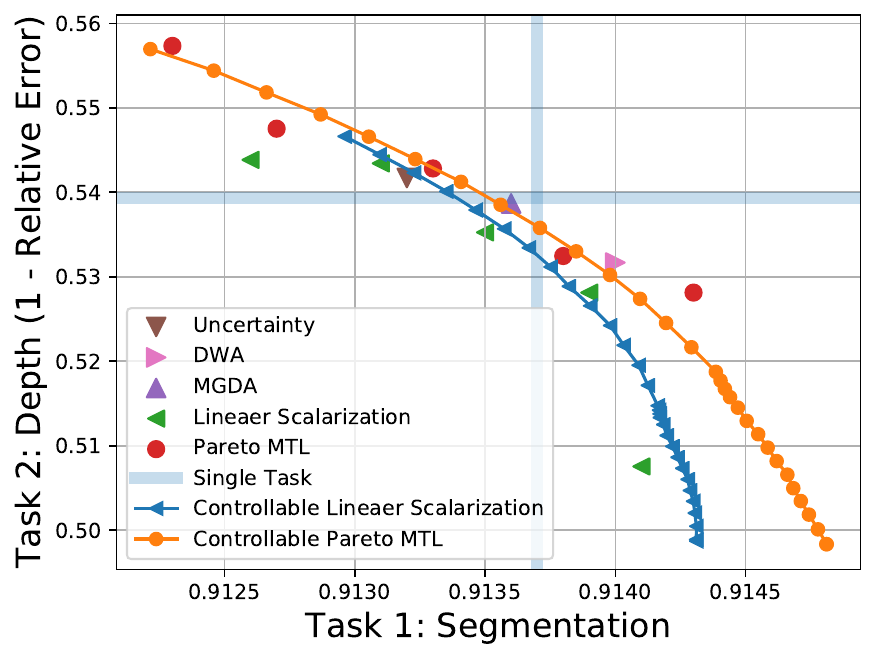}} 
\subfloat[NYUv2]{\includegraphics[width = 0.31\textwidth]{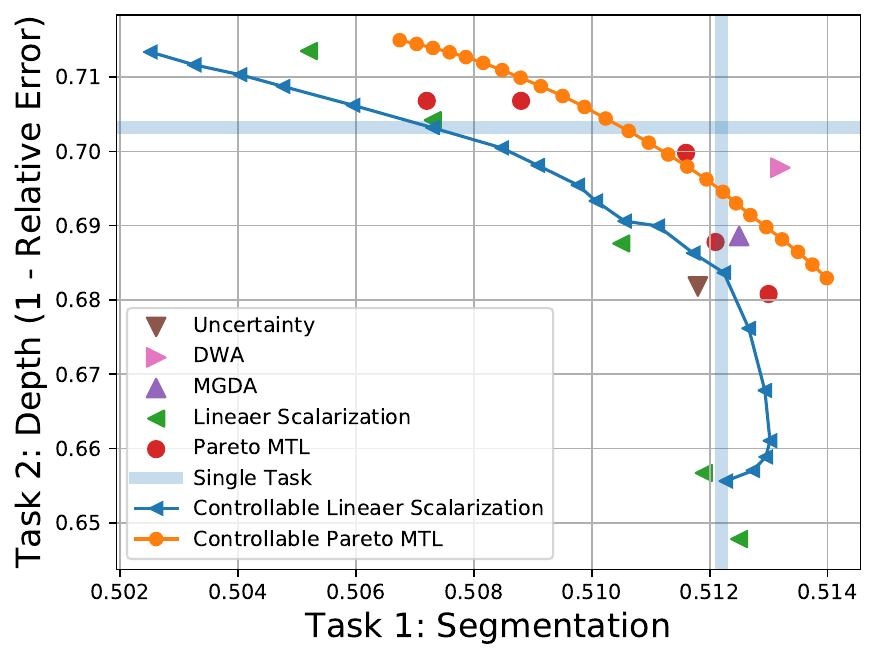}} 
\caption{\textbf{Results on MultiMNIST, CityScapes and NYUv2:} Our proposed algorithm can learn a whole trade-off curve for each problem with a single model respectively. For CityScapes and NYUv2, we report pixel accuracy for segmentation and (1 - relative error) for depth estimation. All problems are to maximize the two objectives. Solutions on the upper right are better than solutions on the lower left. }
\label{exp_multimnist}
\end{figure*}

\begin{figure}[ht]
\centering
\begin{minipage}[c]{0.45\textwidth}
\includegraphics[width = 1\textwidth]{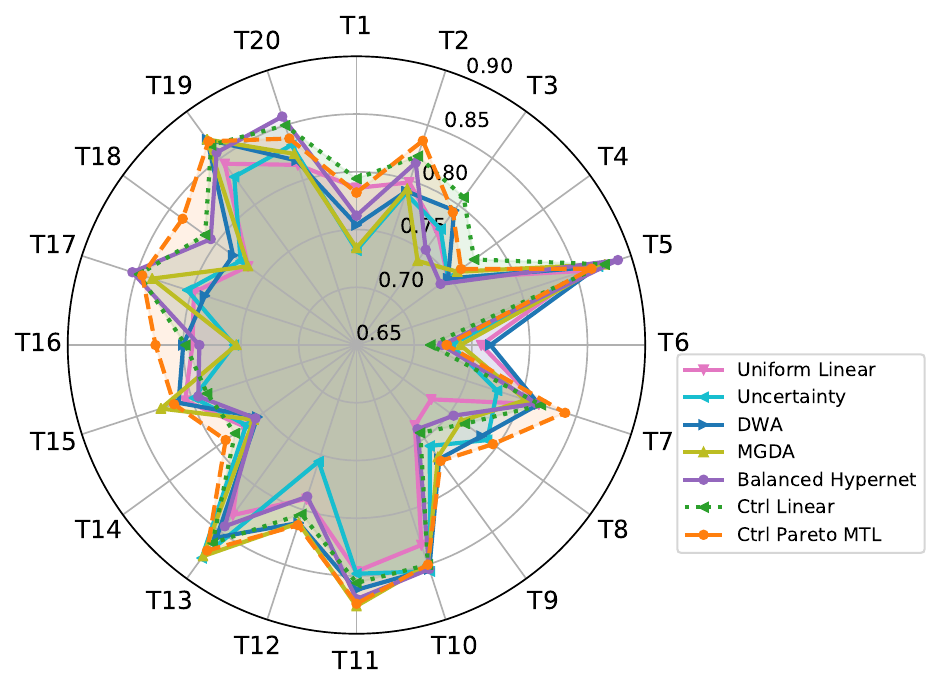}
\end{minipage}
\begin{minipage}[c]{0.40\textwidth}
            \begin{tabular}{cc}
             \hline           & Ave. Acc \\ \hline
            Uniform Linear     & 79.45 $\pm$ 0.23   \\
            Uncertainty     & 79.35  $\pm$ 0.43     \\
            DWA             & 80.52 $\pm$ 0.24       \\
            MGDA         & 80.14  $\pm$ 0.19     \\ \hline
            Balanced Hypernet      & 80.29  $\pm$ 0.13          \\ \hline
            Ctrl Linear     & 81.14 $\pm$ 0.17              \\
            Ctrl Pareto MTL & \textbf{ 81.98 $\pm$ 0.21 }            \\ \hline
            \end{tabular}
\end{minipage}
\caption{\textbf{The results of CIFAR-100 with 20 Tasks:} The results for each method is on a single model respectively. Our proposed methods adjust the preference when making prediction for each task (e.g., preference on task 1 for making prediction on task 1).}
\label{results_many_tasks} 
\end{figure}

In this section, we validate the performance of the proposed controllable Pareto MTL method to generate trade-off curves for different MTL problems. We compare it with the following MTL algorithms: \textbf{1) Linear Scalarization}: simple linear combination of different tasks with fixed weights; \textbf{2) Uncertainty}~\citep{kendall2017multi}: adaptive weight assignments with balanced uncertainty; \textbf{3) DWA}~\citep{liu2019end}: dynamic weight average for the losses; \textbf{4) MGDA}~\citep{sener2018multi}: multiple gradient descent algorithm, to find one Pareto stationary solution; \textbf{5) Pareto MTL}~\citep{lin2019pareto}: to find a set of wildly distributed Pareto stationary solutions; and \textbf{6) Single Task}: the single task learning baseline. 

We conduct experiments on the following widely-used MTL problems: \textbf{1) MultiMNIST}~\citep{sabour2017dynamic}: This problem is to simultaneously classify two digits on one image. \textbf{2) CityScapes}~\citep{cordts2016cityscapes}: This dataset has street-view RGB images, and involves two tasks to be solved, which are pixel-wise semantic segmentation and depth estimation. \textbf{3) NYUv2}~\citep{silberman2012indoor}: This dataset is for indoor scene understanding with two tasks: a $13$-class semantic segmentation and indoor depth estimation.  \textbf{4) CIFAR-100 with 20 Tasks}: Follow a similar setting in~\citep{rosenbaum2018routing}, we split the original CIFAR-100 dataset~\citep{Krizhevsky2009cifar} into 20 five-class classification tasks. An overview of the models we used for each problem can be found in Table.\ref{model_info}, and the details for these problems can be found in the Appendix. 

\textbf{Result Analysis:} The experiment results on MultiMNIST, CityScapes and NYUv2 are shown in Fig.~\ref{exp_multimnist}(a)(b)(c) respectively, where all models are trained from scratch. In all problems, our proposed algorithm can learn the trade-off curve with a single model, while the other methods need to train multiple models and cannot cover the entire curve. In addition, the proposed preference-conditioned multiobjective optimization method can find a better trade-off curve that dominates the curve found by the simple linear scalarization. These results validate the effectiveness of the proposed model and the end-to-end optimization method.

For all experiments, the single-task learning is a strong baseline in performance with a larger model size ($m$ full models). However, it cannot dominate most of the approximated Pareto front learned by our model. Our models can provide diverse optimal trade-offs among tasks (e.g., in the upper left and lower right area) for different problems and support real-time trade-off adjustment. 

The experimental result on the $20$-tasks CIFAR100 classification problem is shown in Fig.~\ref{results_many_tasks}. Our proposed model can achieve the best overall performance by making a real-time preference adjustment for predicting different tasks. It also outperforms the balanced hypernet approach, which has the same hypernetwork-based model but optimizes different tasks with an equal preference. This result shows the advantage of our proposed model on preference-based modeling and real-time preference adjustment for inference.

\begin{table*}[t]
\centering
\caption{ Overveiw of the models we used in different MTL problems.}
\label{model_info}
\begin{tabular}{cccccc}
\hline
Problem                    & Tasks               & Model                      & Network            & Params & Latency(ms) \\ \hline
\multirow{2}{*}{MNIST}     & \multirow{2}{*}{2}  & \multirow{2}{*}{LeNet}     & Single MTL Network & 32K    & 1.36 $\pm$ 0.17   \\
                           &                     &                            & Pareto MTL Network & 34K    & 1.82 $\pm$ 0.31   \\ \hline
\multirow{2}{*}{CityScape} & \multirow{2}{*}{2}  & \multirow{2}{*}{SegNet}    & Single MTL Network & 25M    & 70.6 $\pm$ 0.65   \\
                           &                     &                            & Pareto MTL Network & 26M    & 98.7 $\pm$ 4.89   \\ \hline
\multirow{2}{*}{NYUv2}     & \multirow{2}{*}{2}  & \multirow{2}{*}{SegNet}    & Single MTL Network & 25M    & 175 $\pm$ 13.4    \\
                           &                     &                            & Pareto MTL Network & 26M    & 201 $\pm$ 12.4    \\ \hline
\multirow{2}{*}{CIFAR100}  & \multirow{2}{*}{20} & \multirow{2}{*}{ConvNet}   & Single MTL Network & 1.4M   & 7.01 $\pm$ 0.44   \\
                           &                     &                            & Pareto MTL Network & 1.6M   & 12.4  $\pm$  1.12 \\ \hline
\multirow{2}{*}{NYUv2}     & \multirow{2}{*}{2}  & \multirow{2}{*}{ResNet-50} & Single MTL Network & 56M    & 353 $\pm$ 24.3    \\
                           &                     &                            & Pareto MTL Network & 96M    & 420 $\pm$ 25.2    \\ \hline
\end{tabular}
\end{table*}

\begin{figure}[t]
    	\centering
        \includegraphics[width= 0.85 \linewidth]{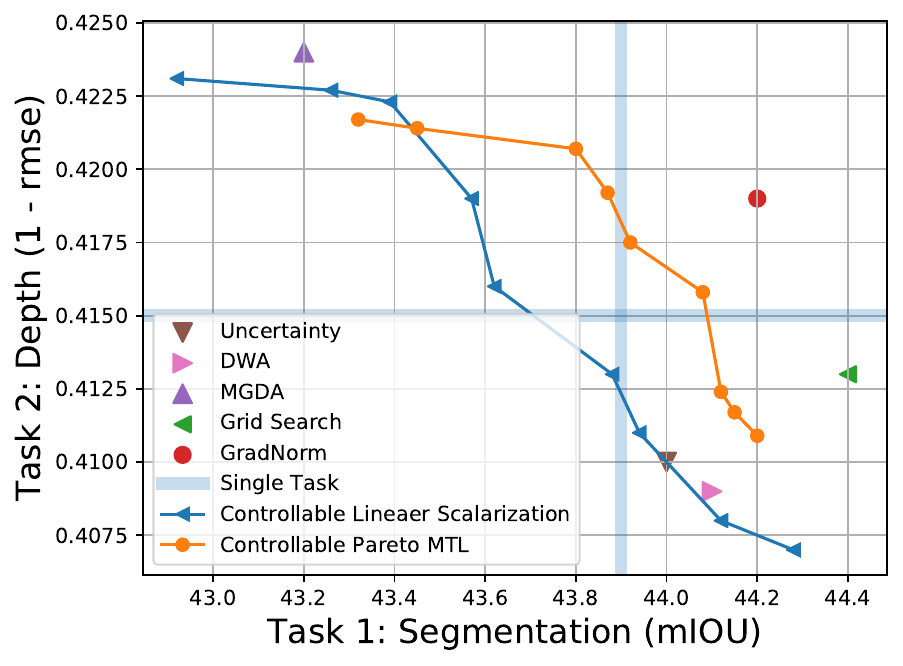}
    	\caption{ \textbf{Results on NYUv2 with ResNet-50 Backbone:} The proposed model can learn reasonable trade-off curves. }
        \label{NYUb2_resnet}
\end{figure}

\textbf{Scalability and Latency:} Table.\ref{model_info} summarizes the information of all models we use in different problems. As discussed in the previous sections, we build the hypernetwork such that the proposed models have a comparable number of parameters with the corresponding single MTL model. Given the current works~\citep{lin2019pareto,mahapatramulti2020multi,ma2020efficient} need to train and store multiple MTL models to approximate the Pareto front, our proposed model is more parameter-efficient while it can learn the whole trade-off curve. In this sense, it is much more scalable than the current methods. The size of hypernetwork-based models can be further reduced due to its compression ability~\cite{ha2017hypernetworks}. Designing more efficient preference-based MTL models is a promising research direction in the future.

The inference latency for all models are also reported. The latency is measured on a 1080Ti GPU with the training batch size for each problem. We report the mean and standard deviation over $100$ independent runs. For our proposed model, we randomly adjust the preference for each batch. Our model has an affordable overhead on the inference latency, which is suitable for real-time preference adjustment. 

\textbf{Experiment with Pretrained Backbone:} We conduct an experiment on the NYUv2 problem with a large-scale ResNet-50 backbone~\citep{he2016deep} following the experimental setting in \citet{vandenhende2020multitask}. Each model has a ResNet-50 backbone pretrained on ImageNet, and task-specific heads with ASPP module~\citep{chen2018encoder}. The results are shown in Fig.\ref{NYUb2_resnet}, where the results for MTL models are from \citet{vandenhende2020multitask} with the best finetuned configurations. Our proposed model also uses the pretrained ResNet-50 backbone, and the hypernetwork generates the encoder's last few layers and all task-specific heads. Our model can learn a good trade-off curve between the two tasks, and its performance could be further improved with task balancing methods and better decoder architecture design as in \citet{vandenhende2020multitask}.

More results and discussions can be found in the Appendix.

\section{Conclusion}
\label{sec_conclusion}

In this paper, we proposed a novel controllable Pareto multi-task learning framework for solving MTL problems. With a preference-based hypernetwork, our method can learn the whole trade-off curve for all tasks with a single model. It allows practitioners to easily make real-time trade-off adjustment among tasks at the inference time. Experimental results on various MTL applications demonstrated the usefulness and efficiency of the proposed method.




\clearpage

\bibliography{icml2021_paper}
\bibliographystyle{icml2021}

\clearpage

\appendix

\section*{Appendix}

We provide more discussion and analysis in this Appendix, which can be summarized as follows:

\begin{itemize}

\item \textbf{Preference-based Multiobjective Gradient:} We give a detailed derivation for preference-based multiobjective gradient descent in Section~\ref{sec_pref_mgda}. 

\item \textbf{Discussion on the Convergence Behavior:} We discuss the convergence behavior of the proposed method in Section~\ref{sec_convergence}. 

\item \textbf{More Experiments:} We compare our proposed methods with the concurrent proposed hypernetwork-based method in Section~\ref{sec_more_experiment}.

\item \textbf{Experimental Setting:} The detailed settings for all experiments are provided in Section~\ref{sec_experiment_setting}.

\item \textbf{Code:} We will make our code publicly available. 

\end{itemize}

\section{Preference-based Multiobjective Gradient Descent}
\label{sec_pref_mgda}

In this section, we give a detailed derivation for the preference-based multiobjective gradient direction and a batched preferences optimization variant for training the hypernetwork-based MTL model. 

\subsection{Preference-based Gradient Direction}

As mentioned in \textbf{Algorithm 1} in the main paper, we use gradient descent to update the solution generator at each iteration. For the case of linear scalarization, calculating the gradient direction is straightforward. In this subsection, we discuss how to obtain a valid gradient direction for the preference-conditioned multiobjective optimization. Similar to the previous work~\citep{sener2018multi,lin2019pareto}, we use multiobjective gradient descent~\citep{fliege2000steepest, desideri2012mutiple} to solve the MTL problem. 

\textbf{The key difference between our proposed method and previous approaches is the parameters to be optimized.} In our algorithm, the optimization parameter is $\vphi$ for the solution generator $\vtheta_{\vp} = g(\vp|\vphi)$ considering all preferences rather than $\vtheta$ for a single solution. 

\begin{figure}[t]
	\centering
\includegraphics[width= 0.80 \linewidth]{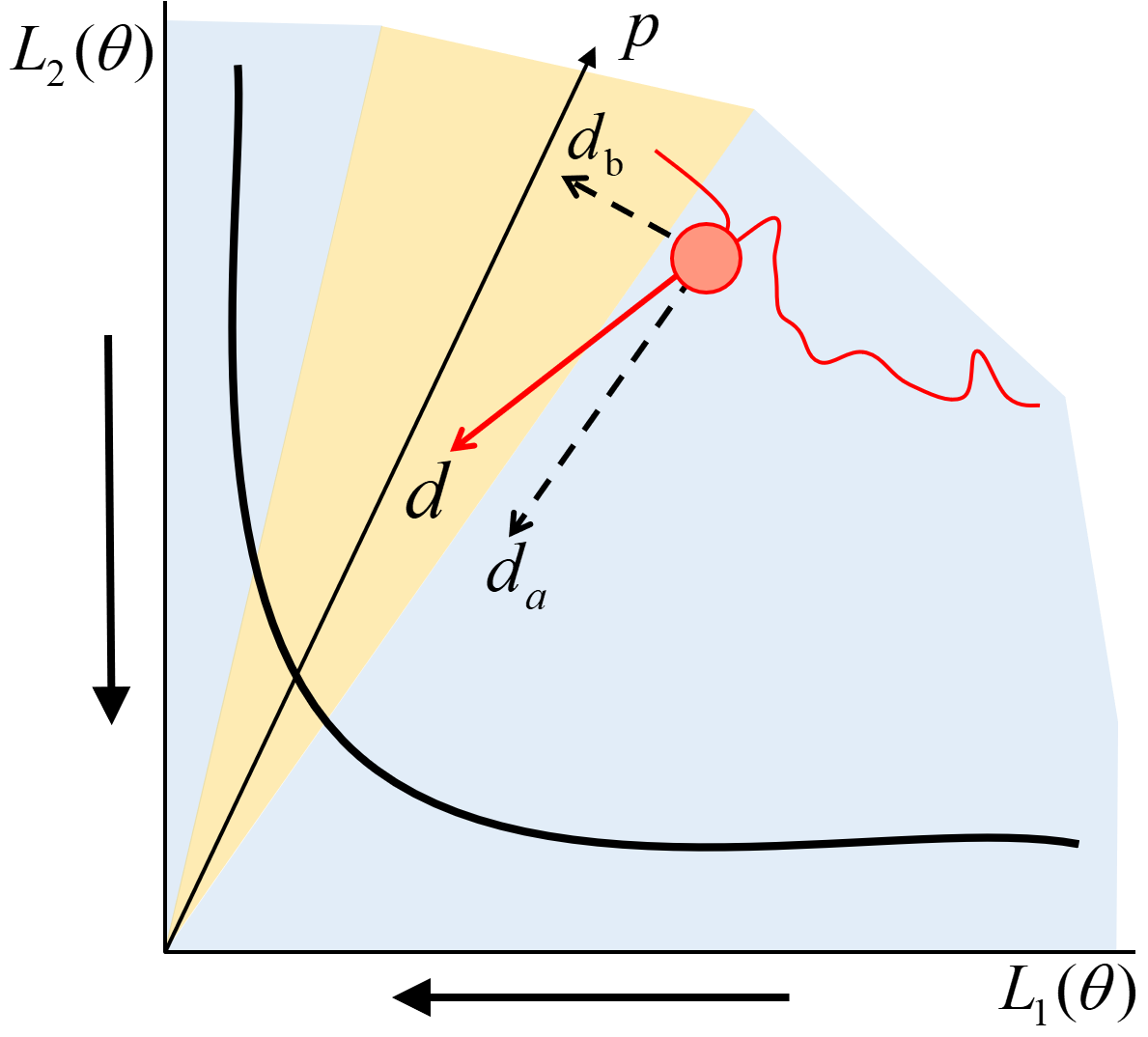}
	\caption{ For a given preference vector, a valid gradient direction should reduce all losses and activated constraints for the corresponding solution. }
    \label{multiobj_gradient}
\end{figure}

A simple illustration of the preference-based multiobjective gradient descent for a problem with two tasks is shown in Fig.~\ref{multiobj_gradient}. At each iteration $t$, the algorithm samples a preference vector $\vp$ and obtains its current corresponding solution $\vtheta_{\vp,t}$. A valid gradient direction should reduce all the losses and guide the generated solution $\vtheta_{\vp,t} = g(\vp|\vphi_t)$ toward the preference region $\Omega(\vp,\vU)$ around the preference vector $\vp$. The preference-based multiobjective optimization problem can be written as:
\begin{eqnarray}
    \label{subregion_app}
    \begin{aligned}
    &\min_{\vphi_t} (\mathcal{L}_1( \vtheta_{\vp,t}),\mathcal{L}_2( \vtheta_{\vp,t}),\cdots, \mathcal{L}_m( \vtheta_{\vp,t})), \\
    &\text{ where}~~  \vtheta_{\vp,t} = g(\vp|\vphi_t) \\
    &\text{ s.t.}~~\mathcal{G}_j( \vtheta_{\vp,t}|\vp,\vU) = (  \vu^{(j)} -\vp)^T \mathcal{L}(\vtheta_{\vp,t}) \leq 0, \\
    & ~\forall j = 1,...,K.
	\end{aligned}
\end{eqnarray}
The generator's parameter $\vphi_t$ is the only trainable parameters to be optimized. With a sampled preference $\vp$, what we need is to find a valid gradient direction $\vd_t$ for updating $\vphi_t$ to reduce all the losses $\mathcal{L}_i(g(\vp|\vphi_t))$ and activated constraints $\mathcal{G}_j(g(\vp|\vphi_t)|\vp,\vU)$.

We follow the methods proposed in~\citep{fliege2000steepest,gebken2017descent}, and calculate a valid descent direction $\vd_t$ by solving the optimization problem:
 \begin{eqnarray}
     \label{submop2}
     \begin{aligned}
        &(\vd_t,\alpha_t) = \text{arg} \min_{\vd\in R^n,\alpha \in R} \alpha + \frac{1}{2} \norm{\vd}^2 \\
        &s.t.~~~ \nabla_{\vphi_t} \mathcal{L}_i(g(\vp|\vphi_t))^T\vd \leq \alpha, i = 1,...,m \\
        & ~~~~~~~~~ \nabla_{\vphi_t} \mathcal{G}_j(g(\vp|\vphi_t))^T\vd \leq \alpha, j \in I(\vphi_t).
     \end{aligned}
 \end{eqnarray}
where $\vd_t$ is the obtained gradient direction, $\alpha$ is an auxiliary parameter for optimization, and  $ I(\theta) = \{j \in I| \mathcal{G}_j(g(\vp|\vphi_t)) \geq 0 \}$ is the index set of all activated constraints. By solving the above problem, the obtained direction $\vd_t$ and parameter $\alpha_t$ will satisfy the following lemma~\citep{gebken2017descent}:

\textbf{Lemma 1:} Let $(\vd_t,\alpha_t)$ be the solution of problem~(\ref{submop2}), we either have:
\begin{enumerate}
     \item A non-zero $\vd_t$ and $\alpha_t$ with
     \begin{eqnarray}\label{eq:bound_2}
    \begin{aligned}
    & \alpha_t \leq -(1/2) \norm{\vd_t}^2 < 0, \\
    & \nabla_{\vphi_t} \mathcal{L}_i(g(\vp|\vphi_t))^T\vd_t \leq \alpha_t, i = 1,...,m \\
    & \nabla_{\vphi_t} \mathcal{G}_j(g(\vp|\vphi_t))^T\vd_t \leq \alpha_t, j \in I(\vphi_t).
    \end{aligned}
    \end{eqnarray}
  \item or $\vd_t = \boldsymbol{0} \in \mathbb{R}^n$, $\alpha_t = 0$, and $\theta_{\vp,t} = g(\vp|\vphi_t)$ is local Pareto critical restricted on $\Omega(\vp,\vU)$.
\end{enumerate}
In case 1, we obtain a valid descent direction $\vd_t \neq \boldsymbol{0}$ which has negative inner products with all $\nabla_{\vphi} \mathcal{L}_i(g(\vp|\vphi_t))$ and $\nabla_{\vphi} \mathcal{G}_j(g(\vp|\vphi_t))$ for $j \in I(\vphi_t)$. With a suitable step size $\eta$, we can update $\vphi_{t+1} = \vphi_{t} + \eta \vd_t$ to reduce all losses and activated constraints. In case 2, we cannot find any nonzero valid descent direction, and obtain $\vd_t = \boldsymbol{0} \in \mathbb{R}^n$, $\alpha_t = 0$. In other words, there is no valid descent direction to simultaneously reduce all losses and activated constraints. Improving the performance for one task would deteriorate the other(s) or violate some constraints. Therefore, the current solution $\vtheta_{\vp} = g(\vp|\vphi_t)$ is a local restricted Pareto optimal solution on $\Omega(\vp, \vU)$.

\subsection{Adaptive Linear Scalarization}

As mentioned in the main paper, we can rewrite the gradient direction $\vd_t$ as a dynamic linear combination with the gradient of all losses $\nabla_{\vphi} \mathcal{L}_i(g(\vp|\vphi_t))$. Similar to the approach in~\citep{fliege2000steepest, lin2019pareto}, we reformulate the problem~(\ref{submop2}) in its dual form:
\begin{align}
\label{submop2_dual}
    &\max_{\lambda_i,\beta_j} -\norm{\sum_{i=1}^{m} \lambda_i
    \nabla_{\vphi_t} \mathcal{L}_i(\vtheta_{\vp,t}) +  \sum_{j \in I(\vphi_t)} \beta_j \nabla_{\vphi_t} \mathcal{G}_j(\vtheta_{\vp,t})}^2 \nonumber, \\
    &\text{ where}~~  \vtheta_{\vp,t} = g(\vp|\vphi_t) \\                    
    &\text{ s.t.}~~~ \sum_{i=1}^{m} \lambda_i + \sum_{j \in I(\vphi_t)} \beta_j = 1, \quad \lambda_i \geq 0, \beta_j \geq 0, \nonumber \\
    &~\forall i = 1,...,m, \forall j \in I(\vphi_t). \nonumber
\end{align}
By solving this problem, we obtain the valid gradient direction $\vd_t = \sum_{i=1}^{m} \lambda_i \nabla_{\vphi_t} \mathcal{L}_i(\vtheta_{\vp,t}) +  \sum_{j} \beta_j \nabla_{\vphi_t} \mathcal{G}_j(\vtheta_{\vp,t})$, where $\lambda_i$ and $\beta_j$ are the Lagrange multipliers for the linear inequality constraints in problem~(\ref{submop2}). 

Based on the definition in problem~(\ref{subregion_app}), we can rewrite the constraint  $\mathcal{G}_j(g(\vp|\vphi_t))$ as a linear combination of all losses:
\begin{eqnarray}
    \label{g_act}
    \begin{aligned}
    \mathcal{G}_j(g(\vp|\vphi_t)) &= (  \vu^{(j)} -\vp)^T \mathcal{L}(g(\vp|\vphi_t)) \\
    &= \sum_{i=1}^{m} (\vu^{(j)}_{i} - \vp_i) \mathcal{L}_i (g(\vp|\vphi_t)).
    \end{aligned}
\end{eqnarray}
Similarly, the gradient of the constraint is also a linear combination of the gradients for all losses:
\begin{eqnarray}
    \label{g_act_grad}
     \begin{aligned}
     \nabla_{\vphi_t} \mathcal{G}_j(g(\vp|\vphi_t)) &= (  \vu^{(j)} -\vp)^T \nabla_{\vphi_t} \mathcal{L}(g(\vp|\vphi_t)) \\
    & = \sum_{i=1}^{m} (\vu^{(j)}_{i} - \vp_i) \nabla_{\vphi_t} \mathcal{L}_i (g(\vp|\vphi_t)).
     \end{aligned}
\end{eqnarray}
Therefore, we can rewrite the valid descent direction as a linear combination of the gradients for all tasks with dynamic weight $\valpha_i(t)$: 
\begin{eqnarray}
    \label{update_mtl_c_app}
    \begin{aligned}
    &\vd_t = \sum_{i = 1}^{m} \valpha_i(t) \nabla_{\vphi_t} \mathcal{L}_{i}(g(\vp|\vphi_t)), \\
    &\text{where }~~\valpha_i(t)  = \lambda_i + \sum_{j \in I_{\epsilon(\theta)}} \beta_j (\vu^{(j)}_{i} - \vp_i).
    \end{aligned}
\end{eqnarray}
The coefficient $\lambda_i$ and $\beta_j$ are obtained by solving the dual problem (\ref{submop2_dual}).  

We use the Frank-Wolfe algorithm~\citep{jaggi2013revisiting} to solve the problem~(\ref{submop2_dual}) as in the previous work~\citep{sener2018multi,lin2019pareto}. We use simple uniform distribution to sample both the unit preference vector $\vp$ and unit reference vectors $\vU$ in this paper. The number of preference vector is $1$ in the previous discussion, and the number of reference vectors is a hyperparameter, which we set it as $3$ for all experiments. 

\subsection{Batched Preferences Update}
\label{sec_batch}

\begin{figure*}[t]
\centering
\subfloat[Iteration 1]{\includegraphics[width = 0.22\textwidth]{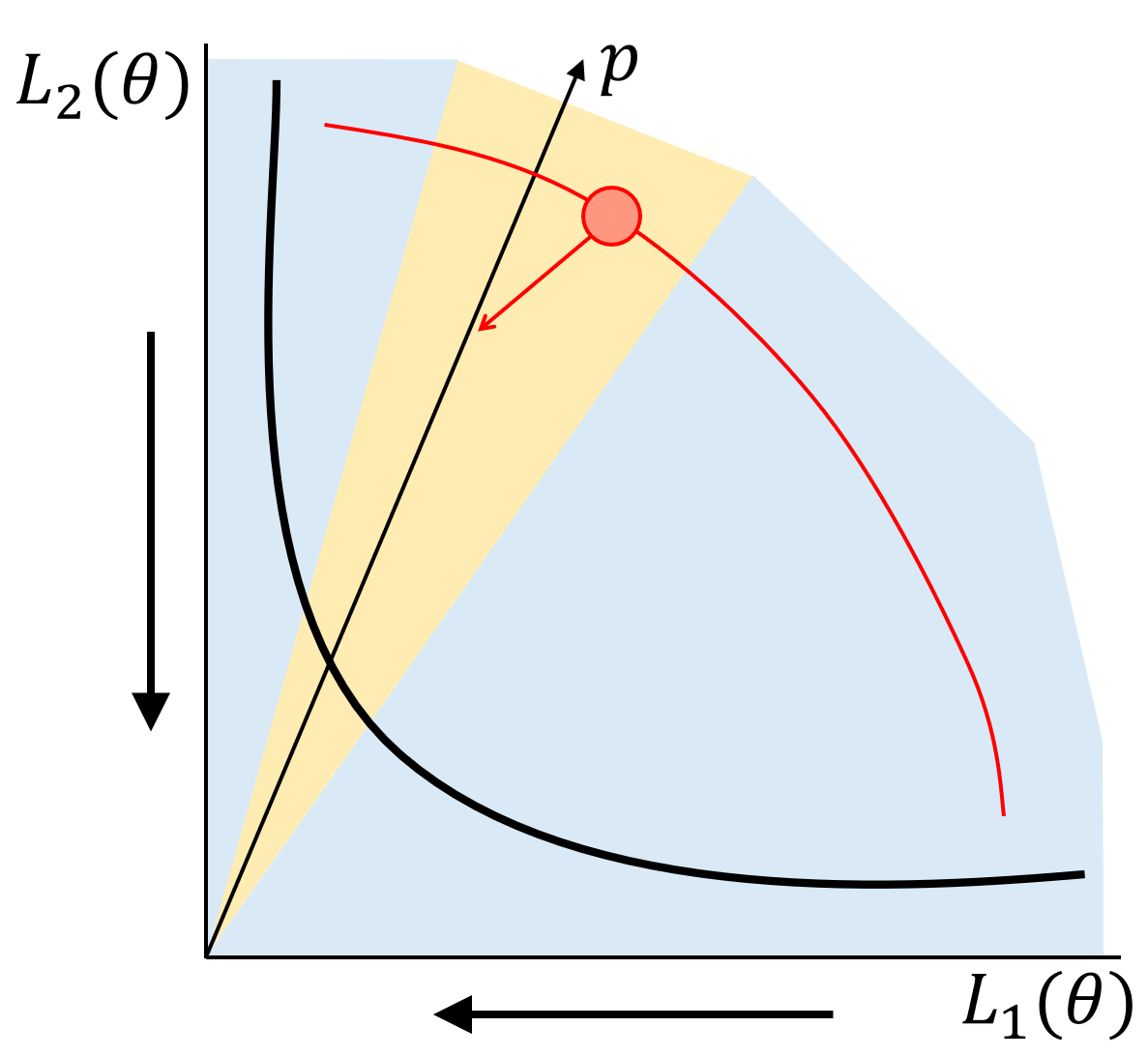}}
\subfloat[Iteration 2]{\includegraphics[width = 0.22\textwidth]{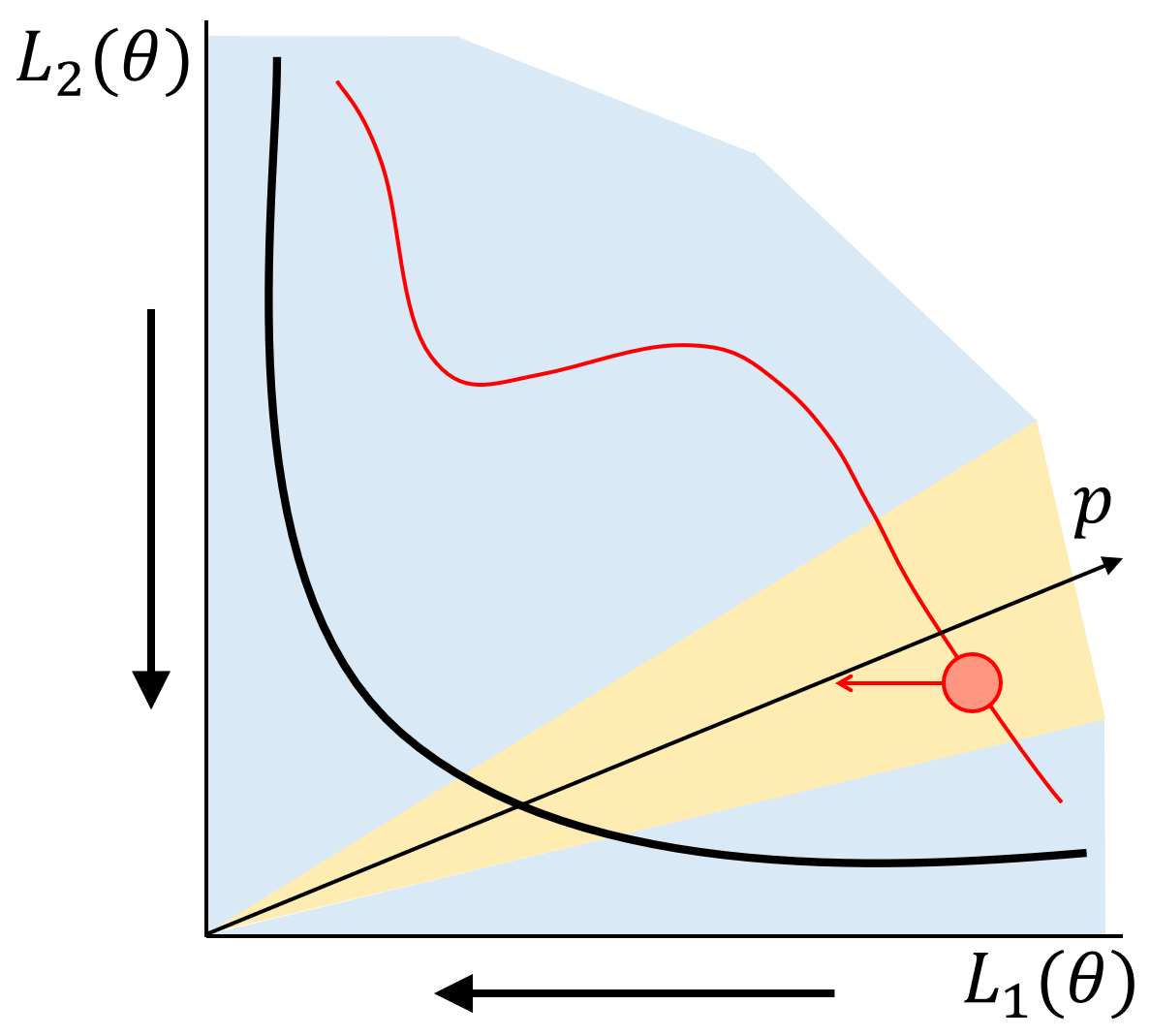}}
\subfloat[Iteration 3]{\includegraphics[width = 0.22\textwidth]{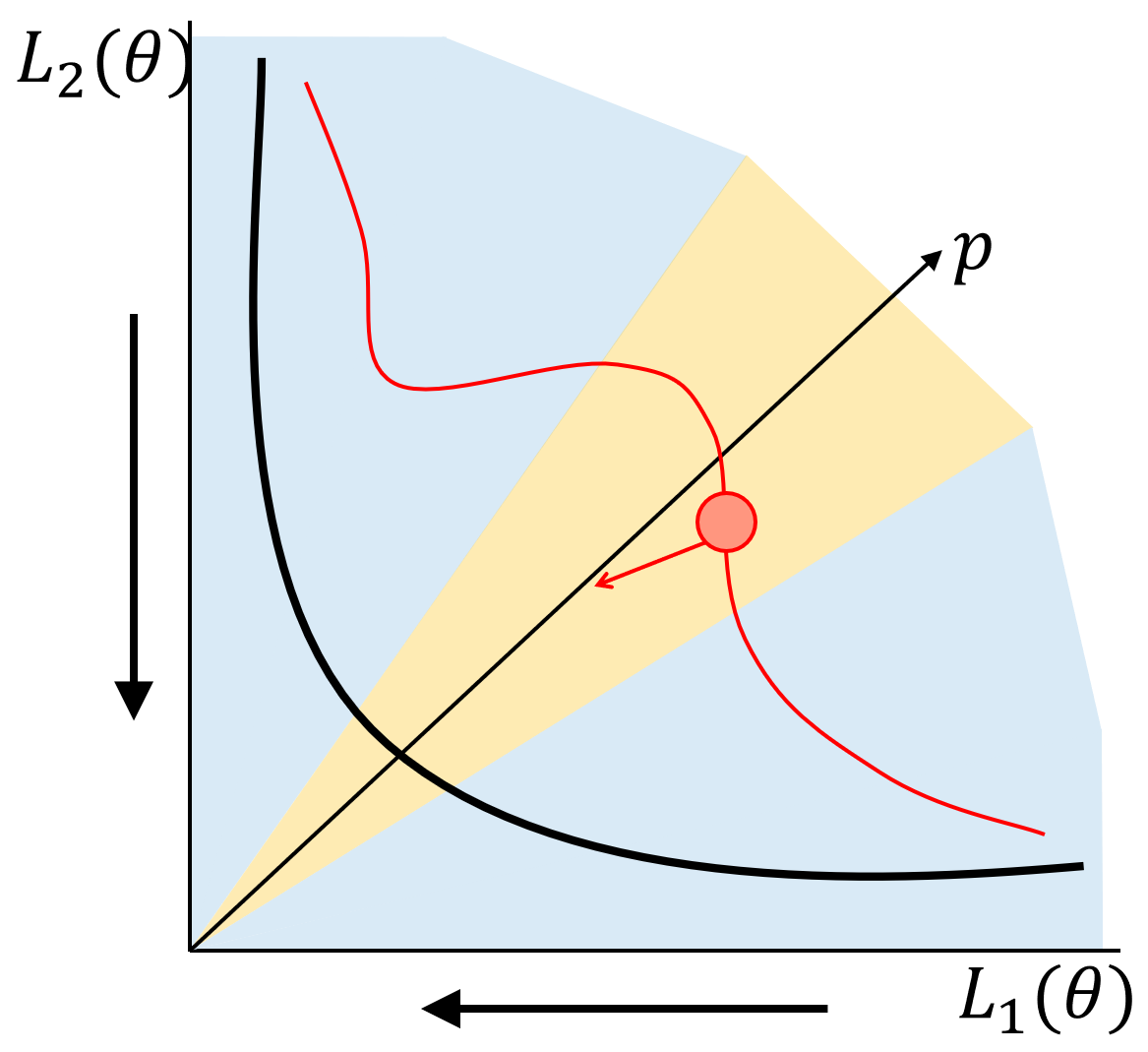}}
\subfloat[Iteration 4]{\includegraphics[width = 0.22\textwidth]{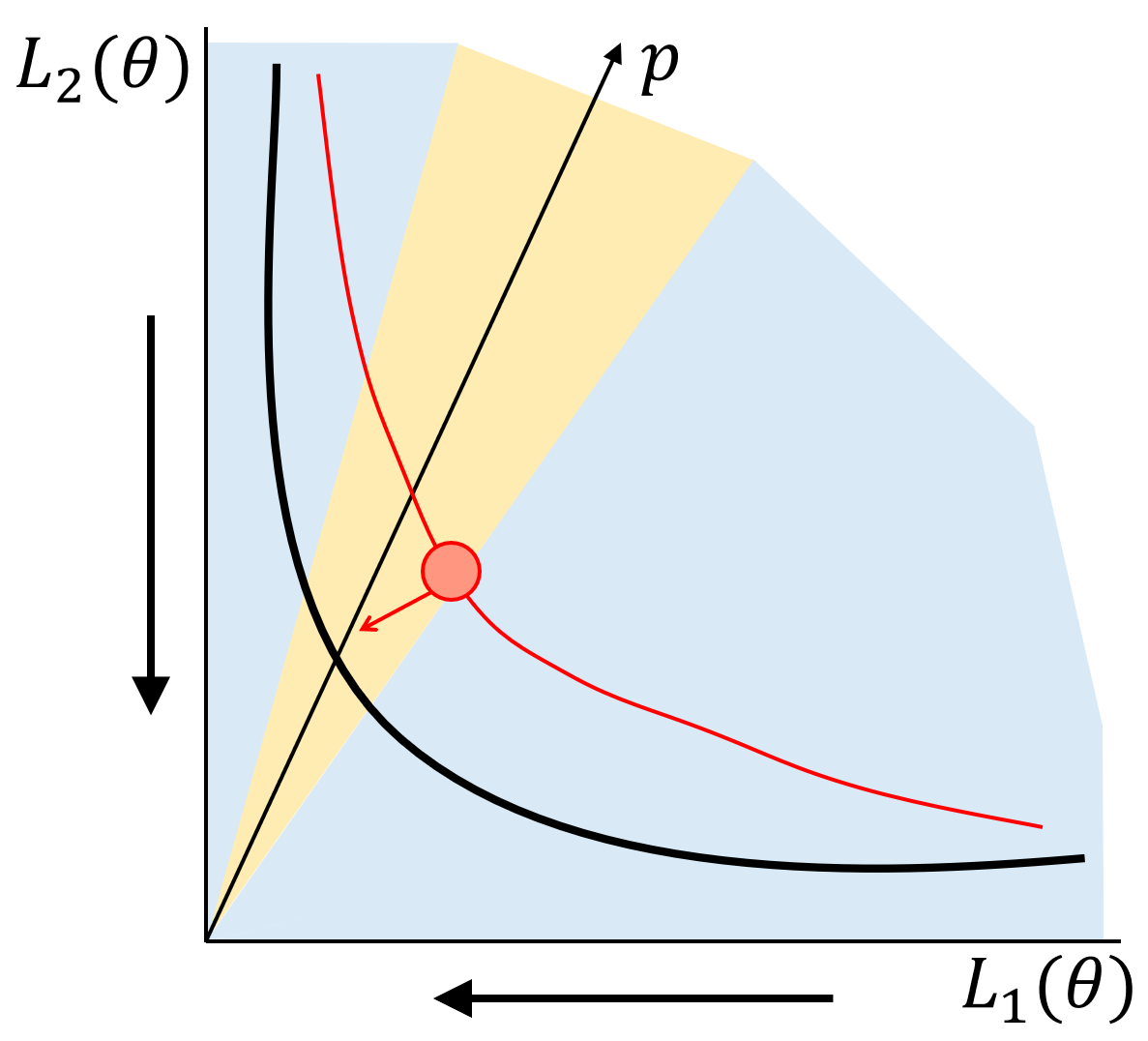}}
\caption{\textbf{The training process for the solution generator.} At each iteration, our proposed algorithm randomly samples a preference vector and calculates a valid gradient direction to update the solution generator. The algorithm iteratively learns and improves the generated manifold during the optimization process.}
\label{fig_mnist}
\end{figure*}

In the main paper, we sample one preference vector at each iteration to calculate a valid direction $\vd_t$ to update the Pareto generator. A simple and straightforward extension is to sample multiple preference vectors to update the Pareto solution generator.

At each iteration, we can simultaneously sample and optimize multiple preferences:
\begin{eqnarray}\label{eq_multi_point_manifold}
 \min_{\vphi_t} \{\mathcal{L}(g(\vp_1|\vphi_t)),\mathcal{L}(g(\vp_2|\vphi_t)),\cdots,\mathcal{L}(g(\vp_K|\vphi_t))\}, 
\end{eqnarray}
where $\vp_1,\vp_2,\cdots,\vp_K$ are $K$ randomly sampled preference vectors and each $\mathcal{L}(g(\vp_k|\vphi_t))$ is a multiobjective optimization problem. Therefore, we now have a hierarchical multiobjective optimization problem. If we do not have a specific preference among the sampled preference vectors, the above problem can be expanded as a problem with $Km$ objectives: 
\begin{eqnarray}
\min_{\vphi_t}
    \left(
    \begin{array}{c} 
    \mathcal{L}_1(g(\vp_1|\vphi_t)), \\ \cdots \\ \mathcal{L}_m(g(\vp_1|\vphi_t)), \\ \cdots \\ \mathcal{L}_1(g(\vp_K|\vphi_t)), \\ \cdots \\ \mathcal{L}_m(g(\vp_K|\vphi_t))
    \end{array}
    \right).
\label{mop_long}
\end{eqnarray}
For the linear scalarization case, the calculation of valid gradient direction at each iteration $t$ is straightforward:
\begin{eqnarray}
\begin{aligned}
 \vd_t &= \sum_{k=1}^{K} \nabla_{\vphi_t} \mathcal{L}(g(\vp_k|\vphi_t)) \\ 
 &= \sum_{k=1}^{K} \sum_{i = 1}^{m} \vp_i \nabla_{\vphi_t} \mathcal{L}_{i}(g(\vp_k|\vphi_t)), 
 \end{aligned}
\end{eqnarray}
where we assume all sampled preferences are equally important.

It is more interesting to deal with the preference-conditioned multiobjective optimization problem. For each preference vector $\vp_k$, suppose we use the rest preference vectors as its reference vectors, the obtained preference-conditioned multiobjective optimization problem would be: 
\begin{eqnarray}
    \begin{aligned}
    \min_{\vphi_t} 
    &(\mathcal{L}_1(g(\vp_1|\vphi_t)),\cdots,\mathcal{L}_m(g(\vp_K|\vphi_t)))  \\
    \text{ s.t.}~~&\mathcal{G}_{1j}(g(\vp_1|\vphi_t)) = (  \vp_j -\vp_1)^T \mathcal{L}(g(\vp_1|\vphi_t)) \leq 0, \\ &\quad\qquad\qquad\qquad\qquad\qquad \forall j \in \{1,...,K\} \backslash \{1\}, \\
    &\mathcal{G}_{2j}(g(\vp_2|\vphi_t)) = (  \vp_j -\vp_2)^T \mathcal{L}(g(\vp_2|\vphi_t)) \leq 0,  \\ 
    &\quad\qquad\qquad\qquad\qquad\qquad \forall j \in \{1,...,K\} \backslash \{2\},  \\
    & \cdots \\
    &\mathcal{G}_{Kj}(g(\vp_K|\vphi_t)) = (  \vp_j -\vp_K)^T \mathcal{L}(g(\vp_K|\vphi_t)) \leq 0, \\
    &\quad\qquad\qquad\qquad\qquad\qquad \forall j \in \{1,...,K\} \backslash \{K\}.
    \end{aligned}
\end{eqnarray}
There are $(K-1)$  constraints for each preference vector, hence total $K(K-1)$ constraints for the preference-conditioned multiobjective problem, although many constraints could be inactivated during training. Similar to the single preference case, we can also calculate the valid gradient direction in the form of adaptive linear combination as:
\begin{eqnarray}
    \label{update_mtl_multipoint}
    \vd_t = \sum_{k=1}^{K} \sum_{i = 1}^{m} \vbeta_i(t)  \nabla_{\vphi_t} \mathcal{L}_{i}(g(\vp_k|\vphi_t)),
\end{eqnarray}
where the adaptive weight $\vbeta_i(t)$ depends on all loss functions $ \mathcal{L}_{i}(g(\vp_k|\vphi_t))$ and activated constraints $\mathcal{G}_{kj}(g(\vp_k|\vphi))$.

\section{Convergence Analysis}
\label{sec_convergence}

We have shown that our proposed method can find good trade-off curves for different large scale MTL problems via a single model. However, similar to other multiobjective optimization algorithms, our method can not guarantee to find the ground-truth Pareto front for a general problem.

As discussed in the main paper and previous section, the general goal for learning the solution generator is:
\begin{eqnarray}
\label{expectation}
 \min_{\vphi} \bbE_{\vp \sim P_{\vp}} \mathcal{L}(g(\vp|\vphi)).
\end{eqnarray}
It is hard to analyze the proposed method's convergence behavior, especially when training a deep MTL problem with complicated optimization landscapes and infinite preferences. In this section, we briefly discuss the case with single, multiple, and infinite preferences. We hope this discussion can lead to a better understanding of our proposed algorithm, and could be useful for potential future work on approximating and learning the whole Pareto front.   

\begin{figure*}[t]
\centering
\subfloat[MultiMNIST]{\includegraphics[width = 0.30\textwidth]{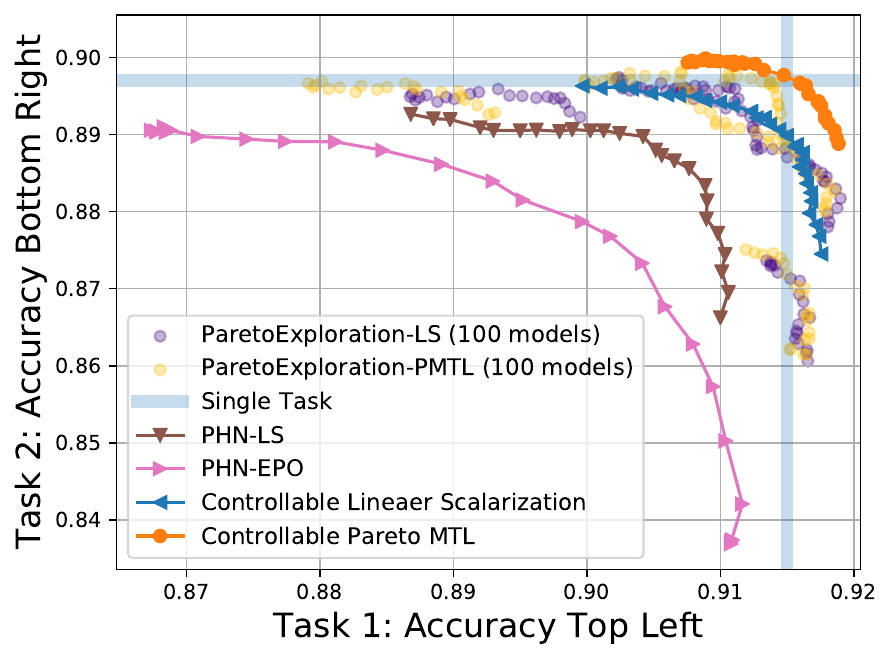}}
\subfloat[MultiFashion]{\includegraphics[width = 0.30\textwidth]{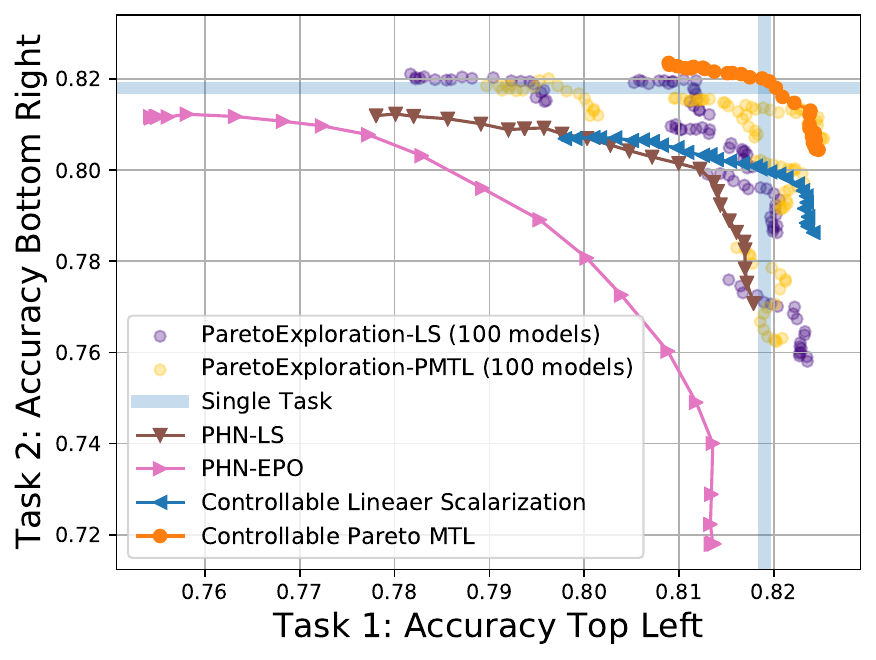}}
\subfloat[MultiFashionandMNIST]{\includegraphics[width = 0.30\textwidth]{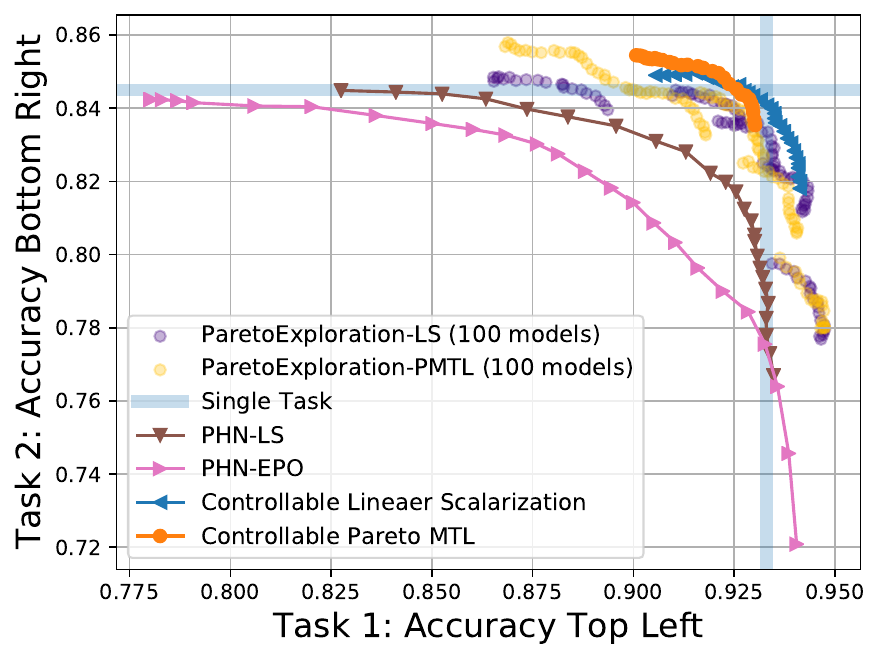}}
\caption{\textbf{The learned trade-off curves on MultiMNIST and its Variants.} We report the testing accuracy for all problems, and solutions on the upper right are better than those on the lower left. Our proposed method can perform better or comparable with other methods, while having a significantly smaller model size. The details are provided in Table.\ref{table_mnist} and Section~\ref{sec_more_experiment}.}
\label{supp_mnist}
\end{figure*}

\textbf{Single Preference.} If we only have a single preference $\vp_s$, the optimization problem will reduce to:
\begin{eqnarray}
 \min_{\vphi} \mathcal{L}(g(\vp_s|\vphi)).
\end{eqnarray}
Since the preference $\vp_s$ is fixed, it is the case for finding a single trade-off solution with a single model as in the previous work~\cite{sener2018multi,lin2019pareto,ma2020efficient,mahapatramulti2020multi}. The only difference is our method has an extra hypernetwork structure.

With a valid gradient direction and proper step size at each iteration as discussed in Section~\ref{sec_pref_mgda}, we can iteratively update $\vphi$ to improve all objective values $ \mathcal{L}(g(\vp_s|\vphi))$. If no valid descent direction can be found, the generated solution $\vtheta_{\vp_s}$ converges to a local Pareto stationary point (might not be the Pareto optimal point for a general problem). Therefore, it shares the same convergence guarantee with the single model counterparts~\cite{sener2018multi,lin2019pareto}.   

\textbf{Fixed and Finite Multiple Preferences.} If we have a set of different but fixed preferences $\{\vp_1,\cdots, \vp_K \}$, the optimization problem would be:
\begin{eqnarray}
 \min_{\vphi_t} \{\mathcal{L}(g(\vp_1|\vphi)),\mathcal{L}(g(\vp_2|\vphi)),\cdots,\mathcal{L}(g(\vp_K|\vphi))\},
\end{eqnarray}
where each $\mathcal{L}(g(\vp_k|\vphi))$ is a single preference-based multiobjective optimization problem as in the previous case. We can use the batched-preference method in Section~\ref{sec_batch} to update the generator for all preferences at each iteration. 

Since different preference-based problems are closely related to each other, their optimal solutions should also share some common properties (e.g., in the same $(m-1)$-dimensional manifold as discussed in the main paper). If the hypernetwork has enough learning capacity, it can generate desired solutions for every single preference-based problem at the same time. With proper assumption, we expect all fixed preferences have the same convergence guarantee with the single-preference case.   

\textbf{Infinite Preferences.} The general and most challenging case is with infinite preferences as in problem~(\ref{expectation}). Since we can not optimize the generator for infinite preferences at one step, we sample one or a set of finite preferences to optimize at each iteration $t$, and have:
\begin{eqnarray}
 \min_{\vphi_t} \mathcal{L}(g(\vp_t|\vphi)), \quad \vp_t \sim P_{\vp}.
\end{eqnarray}
It is the optimization method we use in this paper.

This method is a Monte Carlo sampling and approximation to optimize the expectation, which is similar to stochastic gradient descent (SGD) or batched SGD against full gradient descent. By iteratively sampling and optimizing for different preferences, the solution generator continually learns and improves its performance for all preferences.

As discussed in the main paper, the set of all valid preference vectors $\vp \in P_{\vp}$ is an $(m-1)$-dimensional manifold. Under mild assumptions, the Pareto set is also an $(m-1)$-dimensional manifold, although it is in much higher decision space. With enough learning capacity, we hope the generator can have optimal parameters $\vphi^*$ for all preferences. However, it is challenging to guarantee the generated solutions for infinite preferences are all Pareto stationary points. Even if this is the case, we can not guarantee the generated manifold would well cover the Pareto set and front (even a local one). Future work in this direction would be crucial for designing more efficient methods to learn the whole Pareto front with the convergence guarantee.  

\section{More Experimental Results}
\label{sec_more_experiment}

\begin{table*}[t]
\centering
\caption{ The performance and model sizes of different methods on MultiMNIST problem and its variants.}
\label{table_mnist}
\begin{tabular}{ccccc}
\hline
\multirow{2}{*}{Methods}       & \multicolumn{3}{c}{Hypervolume ($\times 10^{-2}$) $\uparrow$ } & \multirow{2}{*}{Params (K) $\downarrow$} \\
                               & MultiMNIST       & MultiFashion     & MultiFashionMNIST   &                             \\ \hline
ParetoExploration-LS           & 16.49            & 10.40            & 15.32               & 32 * 100 = 3200 (1.1\%)     \\
ParetoExploration-PMTL         & 16.55            & 10.19            & \textbf{15.63}      & 32 * 100 = 3200 (1.1\%)     \\ \hline
PHN-LS                         & 16.09            & 9.90             & 14.86               & 3243 (1.04\%)             \\
PHN-EPO                        & 16.02            & 9.66             & 14.75               & 3243 (1.04\%)             \\ \hline
Controllable LS (ours)         & 16.54            & 9.94             & 15.41               & \textbf{34}                 \\
Controllable Pareto MTL (ours) & \textbf{16.74}   & \textbf{10.49}   & 15.23               & \textbf{34}                 \\ \hline
\end{tabular}
\end{table*}

In the main paper, we have shown that our proposed method can successfully learn the trade-off curves for large scale MTL problems with a single model, and support real-time trade-off control with minimal inference overhead. To our best knowledge, this is the first approach to learn the trade-off curves for large scale MTL problems.  

In this section, we compare its performance with two methods on learning the trade-off curves on small scale problems: \textbf{1) Continuous Pareto Exploration~\cite{ma2020efficient}:} It proposes to use a gradient-based Pareto exploration method to continuously generate and store a dense set of separate Pareto stationary solutions; \textbf{2) Pareto HyperNetwork (PHN)~\cite{navon2021learning}:} This is a concurrent work to our method. It also independently proposes using a hypernetwork to learn the Pareto front, but the current method only works on small scale problems.

\textbf{Problems:} We run experiments on the MultiMNIST problem~\cite{sabour2017dynamic} and two variants, namely MultiFashion and MultiFashionMNIST, as used in the above work~\cite{ma2020efficient,navon2021learning}. These problems are to classify two digits, two fashion items, and one digit with one fashion item in a single image. Details can be found in Section \ref{sec_experiment_setting}. 

\textbf{Models:} We use the open-sourced code for both works to reproduce the results~\footnote{https://github.com/mit-gfx/ContinuousParetoMTL}\footnote{https://github.com/AvivNavon/pareto-hypernetworks}. For the continuous Pareto exploration method, we modify the MTL model to let it have the same structure as in our method and PHN.  We use most default hyperparameters in the code, but carefully fine-tune the exploration step size to let it have a good dense approximation. Follow the setting in their paper, we first generate five seed solutions with linear scalarization (LS) or Pareto multi-task learning (PMTL), and then use Pareto exploration to generate 20 new solutions for each seed solution. Therefore, this method has to train and store $105$ full MTL models in total to approximate the Pareto front.

For the Pareto HyperNetwork method (PHN), we use the default hypernetwork and hyperparameters, which has the same main MTL model structure as in our work. We have tried to fine-tune the hyperparameters and learning strategies, but failed to let it have the same performance with our method. We believe the large hypernetwork size (total $3.2M = 3,200K$ parameters) makes it hard to optimize. Recent work on hypernetwork's initialization method~\citep{chang2020principled} and optimization dynamic~\citep{littwin2020on} would be useful to further improve this method's performance with a large hypernetwork.

\textbf{Result Analysis:} The experiment results are shown in Fig.~\ref{fig_mnist} and Table.\ref{table_mnist}. To compare the quality of approximated Pareto front, we also report the hypervolume indicator~\cite{zitzler1999multiobjective} for all methods on each problem. The hypervolume indicator measures the area between a reference point to a set of solutions. Let $S \subset R^m$ be a set of solutions in the objective space, and $z^*$ be a point dominated by all the points in $S$, the hypervolume $H(S)$ of $S$ is defined as the volume of the set:
\begin{eqnarray}
A=\{z \in R^m | \exists y \in S \mbox{ such that } y \prec z \prec z^{*}\}
\end{eqnarray}
With a fixed reference point, a better approximated Pareto front should have a larger hypervolume value. We use a reference point $(0.5, 0.5)$ for all experiments.

Our proposed method can generate better or comparable trade-off curves for all problems but with a significantly smaller model size (only $1.1\%$ of the compared methods). The small model size also makes our method suitable for real-time trade-off control. Our model can have a significantly smaller size due to the similarity among different preference-based solutions (e.g., in the same $(m-1)$-dimensional manifold). With the model compression method discussed in the main paper, we can share a large amount of parameters among different preferences without decreasing their performance. This property makes it scale well for large MTL models, which is crucial for real-world applications.

\section{Experimental Setting}
\label{sec_experiment_setting}

\paragraph{Synthetic Example:} The synthetic example we use in the main paper is defined as:
\begin{eqnarray}
    \label{toy_example_c4}
    \begin{aligned}
        &\min_{\vtheta} (f_1(\vtheta),f_2(\vtheta)), \text{where} \nonumber \\
        &f_1(\vtheta) = 1 -  \exp{\left(- (\vtheta_1 - 1)^2 - \frac{\sum_{i=2}^{n}[\vtheta_i - \sin(5\vtheta_1)]^2}{n-1}  \right)}, \nonumber \\
        &f_2(\vtheta) = 1 - \exp{\left(- (\vtheta_1 + 1)^2\right)}. \nonumber
    \end{aligned}
\end{eqnarray}
We set $n = 10$ and use a simple two-layer MLP network with 50 hidden units on each layer to generate the Pareto solutions based on the preference vectors.

\paragraph{MultiMNIST~\citep{sabour2017dynamic} and Two Variants:} In this problem, the goal is to classify two overlapped digits in an image at the same time. The size of the original MNIST image is $28 \times 28$. We randomly choose two digits from the MNIST dataset, and move one digit to the upper-left and the other one to the bottom right with up to $4$ pixels. Therefore, the input image is in size $36 \times 36$. MultiFashion and MultiFashionMNIST~\cite{lin2019pareto} are two variants for the MultiMNIST problem, which is to classify two fashion items~\cite{xiao2017fashion}, or one fashion item and one digit in a single image.

Similar to the previous work~\citep{sener2018multi,lin2019pareto}, we use a LeNet-based neural network with two task-specific fully connected layers as the main MTL model. The hypernetwork is a simple MLP. Since the LeNet model is small, we do not use any chunk embedding. We let the hypernetwork-based model have a similar number of parameters with a single MTL model. For all methods, the optimizer is Adam with learning rate lr = $3e^{-4}$, the batch size is $256$, and the number of epochs is $200$.

\paragraph{CityScapes~\citep{cordts2016cityscapes}:} This dataset has street-view RGB images, and involves two tasks to be solved, which are pixel-wise semantic segmentation and depth estimation. We follow the setting used in~\citep{liu2019end}, and resize all images into $128 \times 256$. For the semantic segmentation, the model predicts the $7$ coarser labels for each pixel. We use the L1 loss for the depth estimation. We report the experimental results on the Cityscapes validation set. 

We use the MTL network proposed in~\citep{liu2019end}, which has SegNet~\citep{badrinarayanan2017segnet} as the shared representation encoder, and two task-specific lightweight convolution layers. In our hypernetwork-based model, the hypernetwork contains three 2-layer MLPs with $100$ hidden units on each layer, and most parameters are stored in the parameter tensors for linear projection. The preference embedding and chunk embedding are all $64$-dimensional vectors. We also let the hypernetwork-based model have a similar number of parameters with the single MTL model. For all experiments, we use Adam with learning rate lr = $3e^{-4}$ as the optimizer, and the batch size is $12$. We train the model from scratch with $200$ epochs.

\paragraph{NYUv2~\citep{silberman2012indoor}:} This dataset is for indoor scene understanding with two tasks: a $13$-class semantic segmentation and indoor depth estimation.
Similar to~\citet{liu2019end}, we resize all images into $288 \times 384$. For the training from scratch experiment, we use a similar MTL network and hyperparameter setting as for the CityScapes problem, except the batch size is $8$ in this problem.

We also test the performance on models with pretrained encoder on the NYUv2 Dataset. We follow the setting in the recent MTL survey paper~\citep{vandenhende2020multitask}, all models have a ResNet-50 backbone~\citep{he2016deep} pretrained on ImageNet, and two-specific heads with ASPP module~\citep{chen2018encoder}. In our hypernetwork-based model, the hypernetwork has three different 2-layer MLPs with $200$ hidden unit on each layer. One MLP is for generating shared convolution layers on top of the ResNet backbone, and the other two are for each task-specific head. The shared parameters for backbone are unfrozen, and will be adapted during training. We use $100$-dimensional vectors as the preference and chunking embedding. We use Adam with learning rate lr = $1e^{-4}$ as the optimizer, the batch size is $8$, and the total epoch is $200$.

\paragraph{CIFAR100 with 20 Tasks:} To validate the algorithm performance on MTL problem with many tasks, we split the CIFAR-100 dataset~\citep{Krizhevsky2009cifar} into 20 tasks, where each task is a $5$-class classification problem. Similar setting has been used in the previous work with MTL learning~\citep{rosenbaum2018routing} and continual learning~\citep{oswald2020continual}. 

The MTL neural network has four convolution layers as the shared architecture, and $20$ task-specific FC layers. In our proposed model, we have $1$ MLP as the hypernetwork and the preference and chunking embedding are both $32$-dimensional vectors. The optimizer is Adam with learning rate lr = $3e^{-4}$, the batch size is $128$, and the number of epochs is $200$. We report the test accuracy for all $20$ tasks.

\end{document}